\documentclass[runningheads]{llncs}

\usepackage{eccv}

\usepackage{eccvabbrv}

\usepackage[pdftex]{graphicx}
\usepackage{booktabs}

\usepackage{epsfig}
\usepackage{amsmath}
\usepackage{amssymb}
\usepackage{color}
\usepackage{multirow}
\usepackage{xcolor}
\usepackage{balance}
\usepackage[accsupp]{axessibility}

\usepackage[accsupp]{axessibility}  

\usepackage{hyperref}

\usepackage{orcidlink}

\begin{document}

\title{PAV: Personalized Head Avatar from Unstructured Video Collection} 

\titlerunning{PAV: Personalized Head Avatar from Video Collection}

\author{Akin Caliskan\inst{1}\orcidlink{0000-0003-2918-5603} \and
Berkay Kicanaoglu\inst{1}\orcidlink{0009-0002-5270-6120} \and
Hyeongwoo Kim\inst{2}\orcidlink{0000-0002-2509-8230}}

\authorrunning{A. Caliskan et al.}

\institute{Flawless AI \and Imperial College London}

\maketitle


\begin{abstract}

We propose PAV, Personalized Head Avatar for the synthesis of human faces under arbitrary viewpoints and facial expressions. PAV introduces a method that learns a dynamic deformable neural radiance field (NeRF), in particular from a collection of monocular talking face videos of the same character under various appearance and shape changes. Unlike existing head NeRF methods that are limited to modeling such input videos on a per-appearance basis, our method allows for learning multi-appearance NeRFs, introducing appearance embedding for each input video via learnable latent neural features attached to the underlying geometry. Furthermore, the proposed appearance-conditioned density formulation facilitates the shape variation of the character, such as facial hair and soft tissues, in the radiance field prediction. To the best of our knowledge, our approach is the first dynamic deformable NeRF framework to model appearance and shape variations in a single unified network for multi-appearances of the same subject. We demonstrate experimentally that PAV outperforms the baseline method in terms of visual rendering quality in our quantitative and qualitative studies on various subjects.
\end{abstract}

\begin{figure}[h]
  \centering
  \includegraphics[width=0.9\linewidth]{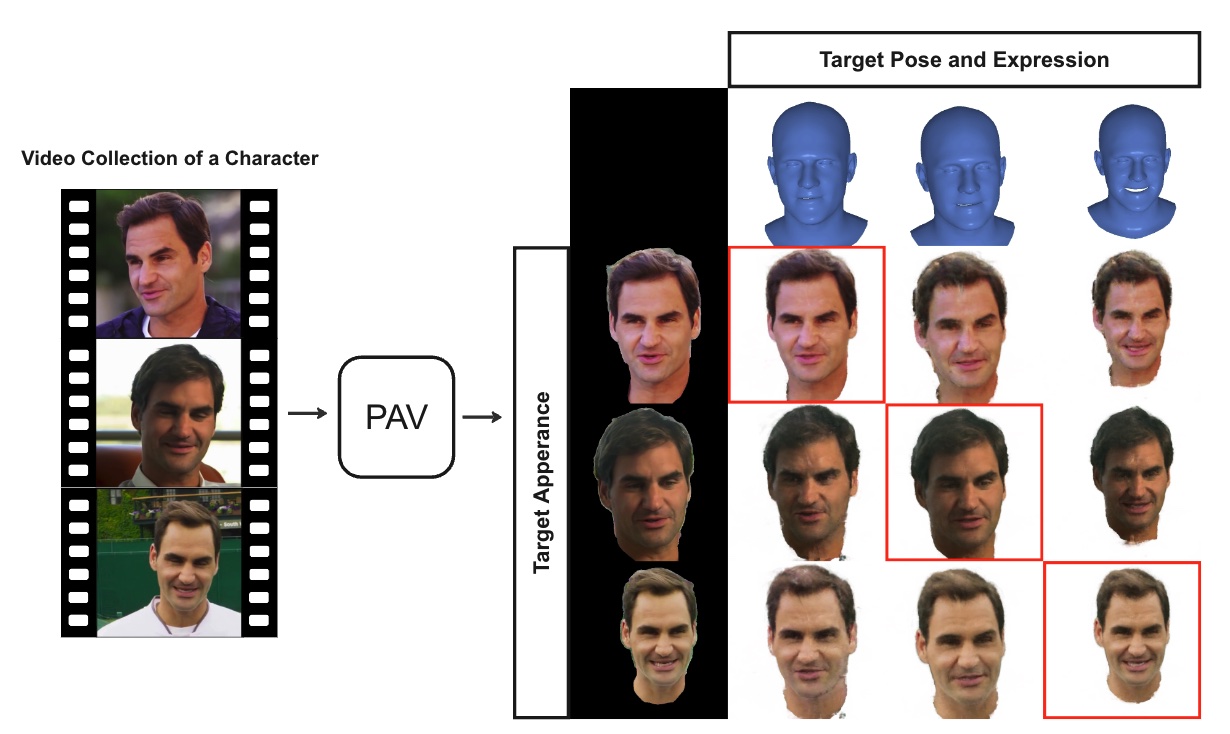}
    \caption{We present an approach for learning personalized head avatars from an unstructured video collection of the same subject. 
    The videos are taken in different environments, with diverse appearances and possible shape changes of a character, as shown on the left.
    Our method leverages all of the input clips into a single unified dynamic deformable NeRF framework and offers high-quality video replays in novel head poses and facial expressions.
    On the right, we show some rendering results of each input appearance of the same subject aligned in head pose and facial expression that are not seen during training. Notably, our method can synthesize identity-specific details while showing coherent expressions across different appearances.}
  \label{fig:teaser}
\end{figure}
\section{Introduction}
\label{sec:intro}


Digital human avatars enabling the editing of facial expressions and head motion have broad applications across telepresence, animation, and digital content creation.
The demands for personalized human avatars such as digital doubles have also increased over the past few years not only in the film industry but in the research community.
To enable the widespread adaptation of these neural avatars, they should be easy to generate and animate under novel poses and facial expressions.
A vast majority of existing methods aim to learn neural avatars from monocular videos to make it more accessible and easy to use \cite{garrido2015vdub, garrido2014automatic, thies2016face2face,kim2018deep, Gafni_2021_CVPR, grassal2022neural, zheng2022avatar, athar2022rignerf, zielonka2023instant}. Early approaches \cite{ garrido2015vdub, garrido2014automatic, thies2016face2face,kim2018deep} mainly rely on 3D morphable models (3DMM) to guide the modeling process and to provide an intuitive control mechanism in novel view synthesis. Despite their success, they do not model the 3D face geometry.
To address this issue, neural radiance fields (NeRFs) \cite{mildenhall2021nerf} have demonstrated volumetric 3D modeling of face geometry and appearance \cite{Gafni_2021_CVPR, grassal2022neural, zheng2022avatar, athar2022rignerf, zielonka2023instant} under novel poses and expressions. While this allows photo-realistic image quality and animation, NeRF-based approaches require per-appearance optimization, maintaining the consistent appearance of the subject, which is constrained to the pre-recording of characters in controlled settings. However, this limits the potential for practical application scenarios where we want to generate a 3D head avatar from a collection of prerecorded video clips of a subject. In this work, we present the Personalized Head Avatar method transforming an unstructured personal video collection, containing images spanning multiple years with different appearances, head poses, and expressions, into a 3D representation of the subject. To create an avatar, we only require an unstructured collection of video clips of the person in arbitrary poses, expressions, and appearances. Our novel framework enables creation of avatars from sparse observations, while yielding high reconstruction quality in novel combinations of pose and appearance, and opens up new applications.

Learning dynamic deformable head NeRF from unstructured video clips is a challenging problem. First, video collections can contain frames of the same identity at different appearances, poses, and expressions. Second, we only have limited pose and expression variation for each appearance, so it is unlikely that all poses and expressions of the face would be well-observed for any given appearance. Given the data, existing NeRF approaches' appearance-specific nature largely limits their potential for practical application scenarios. They require an independent model that is trained from scratch for every appearance from each monocular video. To address these challenges, we model shared deformation field and canonical space neural volumetric representation for all appearances with appearance-dependent latent features. Our key insight is that although observed faces have different appearances across video clips, they should all be explained by a common deformation model with small differences since they all originate from the same person. Furthermore, we assert that every video clip specific to an appearance contains both general and unique information. A single unified network could collaboratively learn the generic information from various appearances, thus this benefits rendering performance as well. The single network is designed to learn a deformation-aware dynamic head NeRF while disentangling appearance-related details from those that are not appearance-specific. For appearance conditioning, we propose employing \textit{learnable appearance embedding} attached to vertices of the tracked mesh and passing them to the dynamic head NeRF network. By adopting this process, PAV is able to learn discriminative latent neural features that are specific for each appearance, resulting in better rendering of appearance-specific details for arbitrary poses and expressions. Unlike existing NeRF methods \cite{Gafni_2021_CVPR, grassal2022neural, zheng2022avatar, athar2022rignerf, zielonka2023instant}, PAV simultaneously trains on all appearances of the same subject, eliminating the need for separate training sessions for each appearance and reducing the overall training duration. Learning of personalized head avatars in a single unified network from an unstructured collection of videos has not been addressed before and is a non-trivial extension of existing monocular head avatar generation methods since they can not handle multiple appearances in a single network. PAV makes this possible through a carefully designed appearance conditioning representation, geometry-attached trainable latent neural texture to learn appearance-specific features, appearance conditioned implicit functions to predict variations in color and density.  We conducted a variety of experiments to show these improvements over the baseline approach. Since there is no such public dataset to evaluate our method, we curate our own in-the-wild video collections for multiple subjects. Overall, we present Personalized Head Avatar with the following contributions:




\begin{itemize}
 \itemsep0em 
    \item We present, PAV, a framework for learning controllable head avatars from unconstrained short video collections where the same person shows different appearances and facial expressions. To the best of our knowledge, this is the first method learning appearance controllable head avatars from in-the-wild monocular video collections of the same subject.
    \item The proposed method can synthesise images of head avatar under target appearance, head pose and facial expressions. 
    \item PAV learns modeling local changes of geometry and color across different appearances of the same identity.
\end{itemize}

\section{Related Work}
\label{sec:related}

\paragraph{Monocular Neural Head Avatars.} 

More recent methods learn neural representations of the person from monocular videos and use neural renderers \cite{tewari2020state,tewari2022advances} to directly generate photo-realistic images in the target head pose and facial expression \cite{averbuch2017bringing,siarohin2019first, mallya2022implicit,Zhang_2023_CVPR,kim2018deep,thies2019deferred,Gafni_2021_CVPR,grassal2022neural,zheng2022avatar,athar2022rignerf,zielonka2023instant,chen2023implicit,gao2022reconstructing,bai2023learning}. These methods are generally classified into image-based or 3D-based neural rendering methods \cite{tewari2020state,tewari2022advances}. The image-based models synthesise the face of a subject without relying on any representation of 3D space. Existing image-based approaches either employ learned warping fields \cite{averbuch2017bringing,siarohin2019first} to deform an input image or encoder-decoder architecture where decoder synthesize the output image \cite{mallya2022implicit,nirkin2019fsgan,wang2018high,zakharov2020fast}. Even though these methods produce high-quality results, they suffer from artifacts for strong head pose and expression changes due to the lack of 3D geometry. The 3D-based neural rendering methods, on the other hand, rely on 3D Morphable Models (3DMM) \cite{egger20203d}. These methods are generally classified into explicit and implicit models. Explicit models utilize 2D neural rendering methods \cite{tewari2020state,tewari2022advances} and rely on neural networks which can be conditioned on coarse RGB renderings based on a linear texture model \cite{kim2018deep}, uv-maps \cite{buhler2021varitex} or latent feature maps \cite{thies2019deferred}. Although these techniques produce high-quality renderings, they are confined to operating within a 2D image space and are limited to craniofacial structures. To tackle this, implicit neural representations for modeling 3D faces have recently become a major focus of attention. In particular, NeRFs \cite{mildenhall2021nerf} shows photorealistic face image synthesis in novel poses and expressions \cite{Gafni_2021_CVPR, grassal2022neural, zheng2022avatar, athar2022rignerf, zielonka2023instant}. Implicit models represent the geometry using implicit surface functions or volumetric representations and they are optimized via volumetric rendering \cite{mildenhall2021nerf}. Early approaches \cite{mildenhall2021nerf} and follow-up works synthesize novel views of complex static scenes using differentiable volumetric rendering. After introduction of NeRF for static scenes, a natural research direction was to generalize it to dyamic and time-varying ones \cite{pumarola2021d,park2021hypernerf}. The reconstruction problem is divided into two different spaces, the deformed space and the canonical space. For human body, a series of approaches have been proposed to target this challenge \cite{peng2021animatable,liu2021neural}. In particular, these methods are differentiated from each other based on deformation field modeling. Gafni et al. \cite{Gafni_2021_CVPR} introduced implicit deformation modeling conditioned on facial expression and head pose. In contrast to modeling the deformation field implicitly, IMAvatar \cite{zheng2022avatar}, RigNerf \cite{athar2022rignerf} and Insta \cite{zielonka2023instant} propose explicit modeling of the deformation field to improve control and rendering quality. Our method falls into this paradigm as we build our system on top of dynamic deformable head NeRF. In particular, we take inspiration from Insta \cite{zielonka2023instant} for designing our framework. However, existing deformable dynamic head NeRF methods including Insta \cite{zielonka2023instant} require per-identity optimization, maintaining the consistent appearance of the same subject. To the best of our knowledge, PAV is the first method to learn a single unified dynamic deformable head NeRF from monocular video of multiple appearances of the same subject.

\paragraph{Geometry-anchored Features for NeRFs.}
PAV introduces appearance-conditioned head NeRF by using geometry-attached features. By using these features, existing works show improved rendering quality for static objects \cite{yang2022neumesh} and human bodies \cite{liu2021neural, bai2023learning}. However, none of the existing methods \cite{liu2021neural,bai2023learning} has yet addressed multi-appearance learning for dynamic neural radiance fields. 




\paragraph{NeRF from Unstructured Data Collections.} 
Learning neural radiance field from unstructured data collections has recently gained much attention. Existing works have enabled photorealistic view synthesis of challenging static scenes, including tourist sites \cite{martin2021nerf} and city-scale scenes \cite{tancik2022block}, and static objects \cite{kuang2022neroic} from internet photo collections. For human subjects, Weng et al. \cite{weng2023personnerf} introduce free-viewpoint human body rendering from an unstructured photo collection. Despite the photo-realistic rendering quality of PersonNerf \cite{weng2023personnerf}, they do not model animatable faces due to the static training images. In our work, we learn animatable 3D head avatars from video collections.

\begin{figure*}
    \centering
    \includegraphics[width=1.0\textwidth]{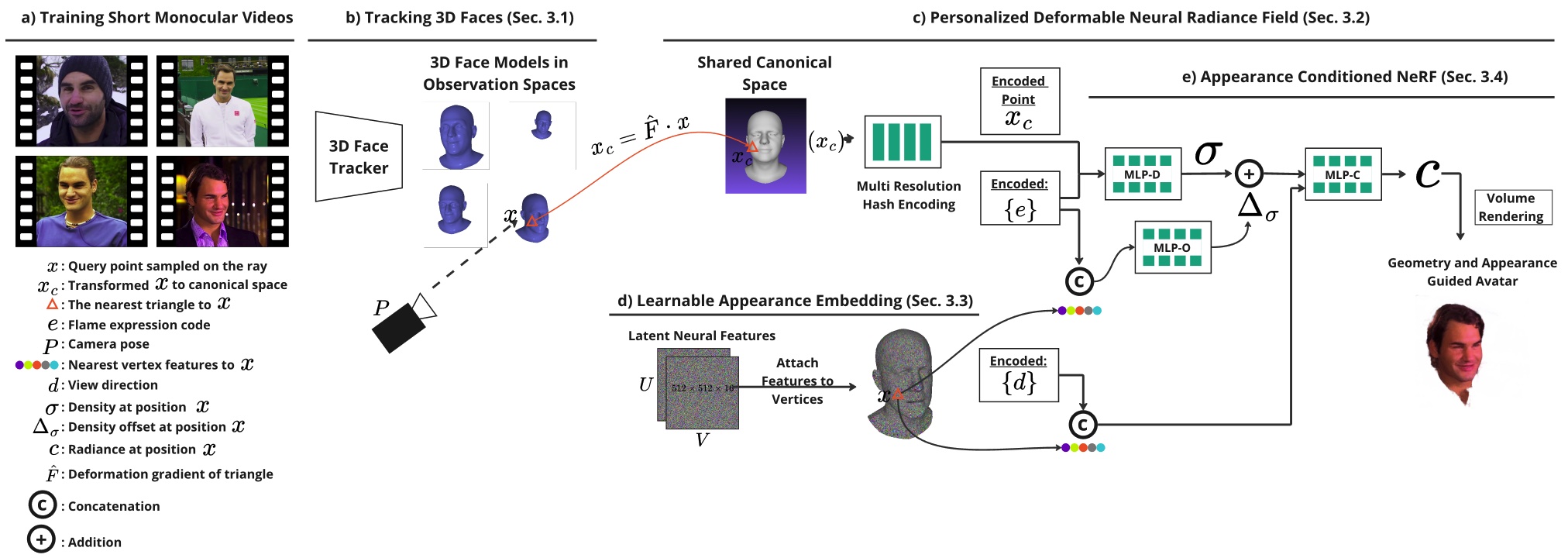}
    \caption
    {Overview of the proposed approach, PAV. (a) shows some training videos in various appearances of the same subject. (b) We estimate the 3D head geometry of the person using FLAME head model. (c) shows the proposed personalized deformable neural radiance field from video collection (d) We learn appearance-specific neural features and attach them to vertices. (d) This feature is passed to fully fused multi-layer perceptrons with additional conditioning on the facial expressions $e$ and the encoded view direction $d$.}
    \label{fig:overview}
\end{figure*}%

\section{Method}
\label{sec:method}

Our goal is to learn a personalized head avatar, PAV, from a monocular video collection of a subject under different appearances, poses, and facial expressions. To create the avatar, we use a small collection $\mathbf{V}{=}\{V_\mathrm{i}\}_{i=1}^N$ with $N$ videos. We parameterize PAV using FLAME \cite{FLAME:SiggraphAsia2017} face model to control head poses, facial expressions and use \textit{neural texture} to model appearance. During inference, we only need the target FLAME parameters and \textit{neural texture}. Learning PAV from a collection of monocular videos requires capturing the geometry and appearance of the dynamic head while also disentangling different appearances of the same identity. In order to tackle this problem, we first capture the dynamic 3D geometry by employing a parametric head model \cite{FLAME:SiggraphAsia2017} (Sec. \ref{sec:head_geometry}). We use the coarse geometry to model geometry-guided deformable neural radiance field (Sec. \ref{sec:pav}) and vertex-attached neural features to condition appearance for rendering (Sec. \ref{sec:app_emb}). We also propose estimating appearance-conditioned neural radiance field, aimed at accurately capturing distinct color and geometry with high fidelity. (Sec. \ref{sec:app_cond_nerf}). An overview of our method can be seen in Fig. \ref{fig:overview}. In the following, we describe each of these modules in greater detail.

\subsection{Head Geometry Estimation} 
\label{sec:head_geometry}

Given the training frames $\mathbf{I}{=}\{I_\mathrm{i}\}$, from all appearances, we first estimate the optimized camera parameters $\mathbf{K} \in \mathbb{R}^{3\times3}$, tracked FLAME \cite{FLAME:SiggraphAsia2017} meshes $\mathbf{M}{=}\{M_\mathrm{i}\}$ with corresponding facial expression coefficients $\mathbf{E}{=}\{E_\mathrm{i}\}$ and poses $\mathbf{P}{=}\{P_\mathrm{i}\}$. We follow an optimization-based method, that minimizes an objective function with photo-consistency and landmark terms \cite{zielonka2022towards}. Different from the original work \cite{zielonka2022towards} which is designed for a single appearance, in PAV, we compute the average shape parameters from all appearances of the same subject to represent identity shape during optimization.

\subsection{Personalized Deformable NeRF}
\label{sec:pav}

Our goal is to create a head avatar that can be learned from frames in video collection of various appearance of the same subject. We parameterize PAV using FLAME model and learned appearance conditioning. Specifically, PAV takes target \textit{head pose} $\theta$, \textit{facial expression} $e$ and \textit{latent neural features (LNF)} $Z_{j} \in \mathbb{R}^{512 \times 512 \times 16}$ in the  texture atlas space (UV space) for an appearance $j$ as inputs and synthesizes the target image $I^{\theta, e, Z_j}$ as:
\begin{align}
I^{\theta, e, Z_j} = \texttt{PAV}(\theta, e, Z_j, K),
\end{align}
where $K$ corresponds to the intrinsic parameters of the target camera viewpoint. For this purpose, we are using a geometry-guided dynamic deformable neural radiance field. However, deformable head NeRF models cannot disentangle appearance from the neural radiance field (See Sec. \ref{sec:related}). To tackle this problem, we employ shared canonical space for all appearances where a shared neural radiance field is constructed. With this shared radiance field at canonical space, to render the target head pose and facial expression using volumetric rendering, we canonicalize the sample points $x$ on a ray from the observation space to query points $x_c$ in the shared canonical space radiance field by using deformation gradient $\hat{F}$: $x_c = \hat{F} \cdot x$ where $ \hat{F} \in \mathbb{R}^{4\times4}$ and $\{x,{x_c}\} \in \mathbb{R}^{4}$. We employ the nearest triangle search for a given point to compute the deformation gradient. Specifically, the deformation gradient is computed between the deformed triangle $\hat{T}_{def} \in M_i$ in the observed mesh and the canonical triangle $\hat{T}_{canon} \in M^{canon}$ in the canonical mesh.
To increase the speed of nearest triangle search, we employ \textit{bounding volume hierarchy (BVH)} \cite{clark1976hierarchical}. We refer readers to \cite{zielonka2023instant} for a detailed description of deformation gradient computation. 

After canonicalization, given the predicted appearance-conditioned radiance $c$ and density $\sigma$ (Sec. \ref{sec:app_cond_nerf}) for every point on each ray, the representation of the avatar is optimized by using the differentiable volumetric rendering. For each camera ray $r(t) = o + td$ with camera center $o$ and viewing direction $d$, the color of the corresponding pixel can be computed by accumulating the predicting radiance and densities of the sample points along the ray: 

\begin{equation} \label{eqn:volume_render}
\mathcal{C}(\mathbf{r}) = \int_{t_n}^{t_f} \sigma(\mathbf{r}(t)) \cdot \mathbf{c}(\mathbf{r}(t)) \cdot \mathcal{T}(t) \,dt
\end{equation} 

\noindent where $t_n$ and $t_f$ are the near and far bounds correspondingly, and $\mathcal{T}(t_n) = \exp{( - \int_{t_n}^{t_f} \sigma(t) \,dt)}$ is the accumulated transmittance along the ray from $t_n$ to $t_f$, $\sigma(t)$ is the density and $c(t)$ it the radiance at position $x_t$. 


\subsection{Learned Appearance Embedding}
\label{sec:app_emb}

In this section, PAV introduces a learnable appearance embedding that is essential and effective for enabling discriminative multi-appearance rendering using single unified network. To achieve this goal (See Fig. \ref{fig:overview}), a set of $N$ learnable latent features $\mathbf{Z}{=}\{Z_\mathrm{i}\}_{i=1}^N, Z_{i} \in \mathbb{R}^{512 \times 512 \times 16}$ are defined in UV space corresponding to $N$ appearances of the same subject in the video collection. We impose \textit{high-dimensional} learnable latent features to condition implicit representations for both density $\sigma$ and radiance $c$ values in the canonical space. To this end, these latent neural features are attached to the vertices of 3DMM mesh, which enables PAV to memorize local variations. For each point $x$ sampled on a ray, we find the nearest triangle $\mathcal{T}{=}\{v_\mathrm{i}\}_{i=1}^3, v_{i} \in \mathbb{R}^{3}$ (sec. \ref{sec:pav}) and aggregate the features $z(v_i)$ of vertices $v_i$ from the vertices of the nearest triangle to get the appearance embedding $Z^x$. Then we directly use the appearance embedding to condition for appearance-dependent neural radiance field estimation (Sec. \ref{sec:app_cond_nerf}). During training, these features are randomly initialized and optimized jointly with the objectives of the entire network.




\subsection{Appearance Conditioned NeRF}
\label{sec:app_cond_nerf}

To model the appearance-conditioned dynamic deformable head NeRF, PAV incorporates learned appearance-embedding (Sec. \ref{sec:app_emb}) into neural radiance field in canonical space (See Fig. \ref{fig:overview}). For a point $x$ sampled on a ray, after canonicalization, the point $x_c$ is encoded using multi-resolution hash encoding \cite{muller2022instant}. This feature $\phi(x_c)$ is passed to multi-layer perceptron for geometric feature estimation. Following \cite{Gafni_2021_CVPR} and \cite{zielonka2023instant}, we conditioned every point features $\phi(x_c)$ with the facial expressions $e$ of frame $i$. Without per-frame latent code, for shared volume density, we learn implicit representation $\mathcal{F}_{density}$ using multi-layer perceptron  (MLP-D in Fig. \ref{fig:overview}):

\begin{align}
\mathcal{F}_{density}: (\phi(x_c),e_i) \rightarrow \sigma
\end{align}

\noindent A primary challenge lies in learning unique density estimations for each appearance while separating the shared density aspects from the appearance.
Hence we design PAV such that appearance-based density offset is predicted from appearance embedding $Z^{x}$ sampled from the nearest triangle to the point $x$ on the mesh $M_i$ and facial expression $e_i$ of frame $i$. For density offset, we learn implicit representation $\mathcal{F}_{offset}$ using multi-layer perceptron  (MLP-O in Fig. \ref{fig:overview}): 
\begin{align}
\mathcal{F}_{offset}: (Z^{x},e_i) \rightarrow \Delta_{\sigma}
\end{align}
Final density value $\sigma(x) = \sigma + \Delta_{\sigma}$ for the point $x$ is the summation of shared density features $\sigma$ and density offset $\Delta_{\sigma}$. For emitted radiance at point $x$, we learn implicit representation $\mathcal{F}_{color}$ using multi-layer perceptron (MLP-C in Fig. \ref{fig:overview}): 
\begin{align}
\mathcal{F}_{color}: (Z^{x},d_i,\sigma(x)) \rightarrow c
\end{align}
\noindent where $d_i$ is the encoded view direction of training frame $i$. Then, final color for the pixel is computed by using differentiable \textit{volume rendering} (Eq. \ref{eqn:volume_render})





\subsection{Optimization}
Our model is trained on a collection of monocular RGB videos with photometric and depth losses. For photometric loss $\mathcal{L}_{color}$, we employ $L_2$ distance between the renderings $C$ and ground-truth images $I$, $\mathcal{L}_{color} = \sum_{i}^{} \sum_{r}^{} \|C_i(r) - I_i(r)\|$, where $r$ denotes the camera ray of each pixel and $i$ denotes frame index. For depth loss $\mathcal{L}_{depth}$, we rasterize the depth of tracking mesh $M_i$ and apply $L_1$ distance between this map $D$ and the ray termination $H$ of the volumetric rendering only for face region as FLAME model is restricted to model details like hair: $\mathcal{L}_{depth} = \sum_{i}^{} \sum_{r}^{} \mathcal{M}_i(r)\|H_i(r) - D_i(r)\|$, where $\mathcal{M}$ denotes face binary mask. The total loss is defined as: 
\begin{equation} \label{eq:overall_loss}
\mathcal{L} =  \lambda_{color}{\mathcal{L}_{color}} + \lambda_{depth}{\mathcal{L}_{depth}}
\end{equation}

\noindent where we emprically set $\lambda_{color} = 50$ and $\lambda_{depth} = 10$. 



\section{Experiments}

In this section, we evaluate the performance of PAV using two different datasets. We perform an ablation study to validate our design choices and a comparison against the state-of-the-art monocular head avatar method in various configurations.

\subsection{Dataset}
\label{sec:exp_dataset}

Our method takes a video collection of a subject containing different appearances, poses, and expressions. Since there is no existing dataset to evaluate learning 3D head avatar from video collections of the same subjects with various appearances, poses, and expressions, we compiled a new Video Collection \textit{(VidCol)} dataset from talking face videos of \textit{Roger Federer},\textit{King Charles} and \textit{Tom Hanks} across years. This dataset consists of in-the-wild videos recorded in uncontrolled environments. In VidCol dataset, we curated videos from 5 different appearances for each subject. We randomly sampled $500$ frames at $512^2$ resolution from each video (or appearance) and we used a total of $2500$ frames to train and test a single unified network per subject. We will share this dataset with annotations of semantic masks, tracked 3DMM models, and camera calibrations for further research in this direction. 

For our experiments, we evaluate all characters and train a separate PAV model for each subject. We develop two different protocols for evaluation:

\paragraph{a) Unified Network for Novel Pose/Exp. Synthesis} This protocol evaluates the performance in terms of novel appearance, head pose, and facial expression synthesis when training a single unified network from all appearances. For training, we select 400 frames for each appearance of a subject talking in front of a camera with various head poses. It mimics a training video commonly used for monocular neural head avatar methods (Sec. \ref{sec:related}). For testing, we use the remaining 100 frames for each appearance with unseen poses and expressions. 

\paragraph{b) Separated Networks for Novel Pose/Exp. Synthesis} This protocol evaluates the novel appearance, head pose, and facial expression generalization ability of the methods when training a separate network for each appearance. We select 400 training images for each appearance of a subject talking of a camera, same as protocol-a. For testing, we choose the remaining 100 unseen pose/expression combinations with ground truth, same as protocol-a, and another 300 unseen pose/expression sets without ground truth image data. 







\subsection{Metrics}

We report several metrics to evaluate the quality of synthesized images as well as the ablation study of our method. For synthesized images with provided ground-truth, we use Learned Perceptual Patch Similarity (LPIPS) \cite{zhang2018perceptual}, Deep Image Structure and Texture Similarity (DISTS) \cite{ding2022dists}, Structural Similarity Index (SSIM) \cite{wang2004ssim}, Peak Signal-to-Noise Ratio (PSNR) and Fréchet Inception Distance (FID) \cite{heusel2017gans}. When there is an absence of ground truth, we adopt FID metric for quantitative comparison.

\subsection{Implementation Details}

We optimize Eq. \ref{eq:overall_loss} using Adam optimizer \cite{kingma2014adam} with hyperparameters $\beta_1=0.9$ and $\beta_2=0.99$. We set the learning rate to $2.5 \times 10^{-3}$ for $F_{density}$, $F_{color}$ and $Z$, and to $2.5 \times 10^{-4}$ for $F_{offset}$. We sample 1024 points along each ray for rendering. Multi-resolution hash encoding used in the method is set to 16 \textit{number of levels}, 18 \textit{hash map size}, and 2048 \textit{finest resolution}. The optimization takes 
15K iterations to converge when training network with all appearances of single subjects. More details on the network are provided in the supplementary document. 



\begin{figure*}[!t]
  \centering
  \small
    \setlength{\tabcolsep}{1pt}
    \renewcommand{\arraystretch}{0.0}
\scalebox{1.0}{
  \begin{tabular}{cccccc}
    \includegraphics[width=0.16\linewidth]{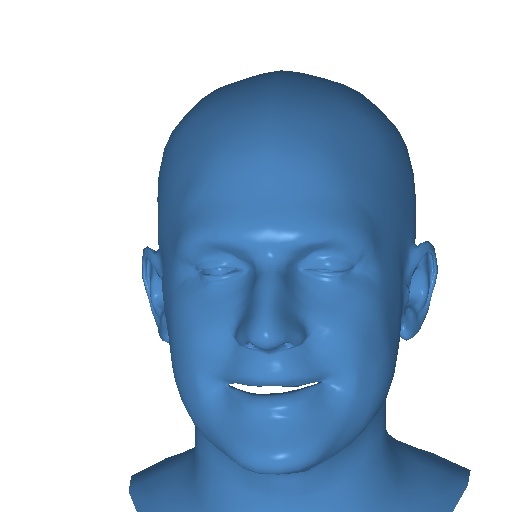} &
    \includegraphics[width=0.16\linewidth]{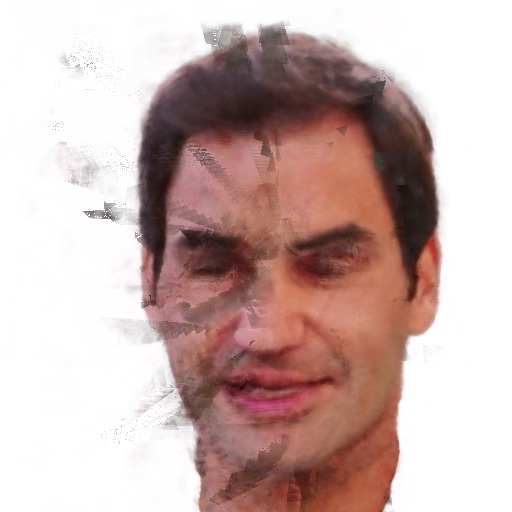} &
    \includegraphics[width=0.16\linewidth]{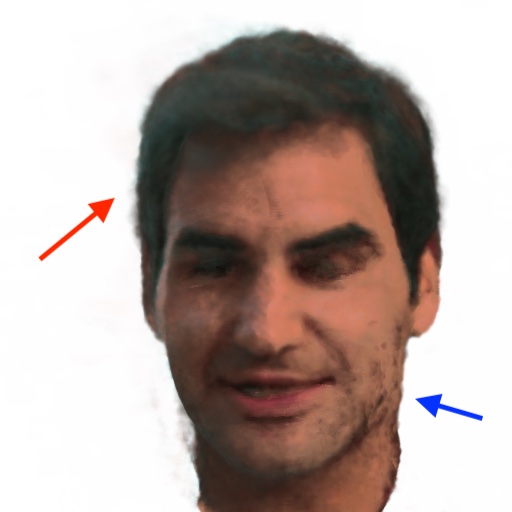} &
    \includegraphics[width=0.16\linewidth]{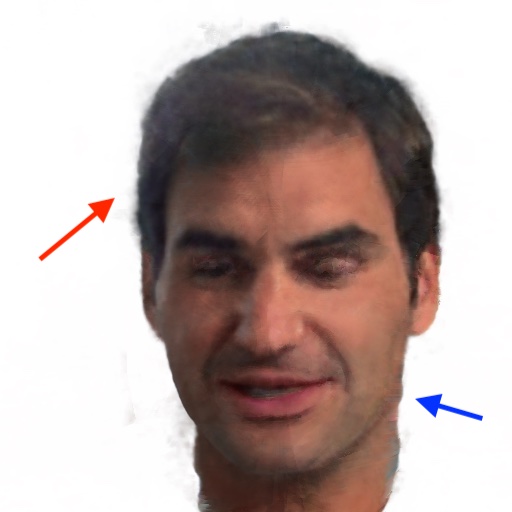} &
    \includegraphics[width=0.16\linewidth]{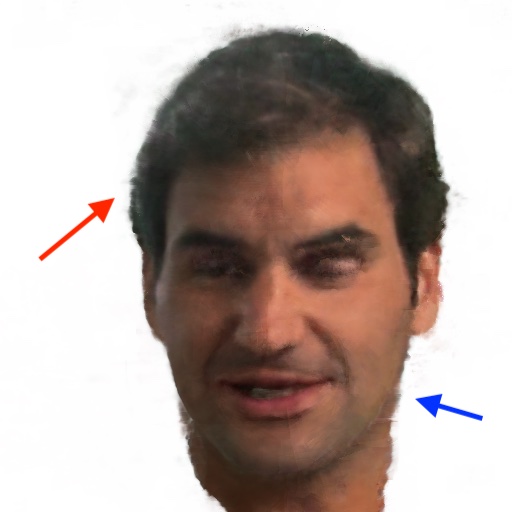} &
    \includegraphics[width=0.16\linewidth]{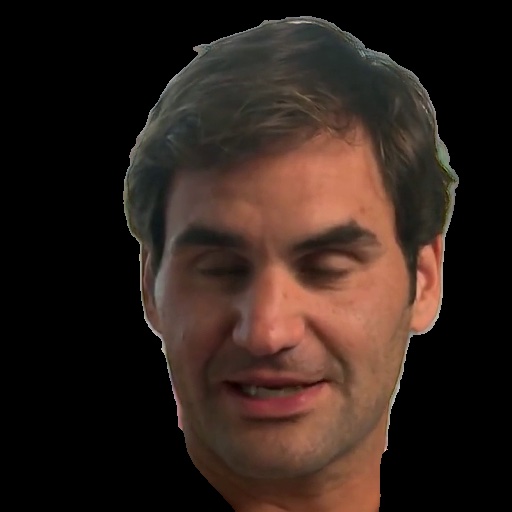}
      \\
       &
    \includegraphics[width=0.16\linewidth]{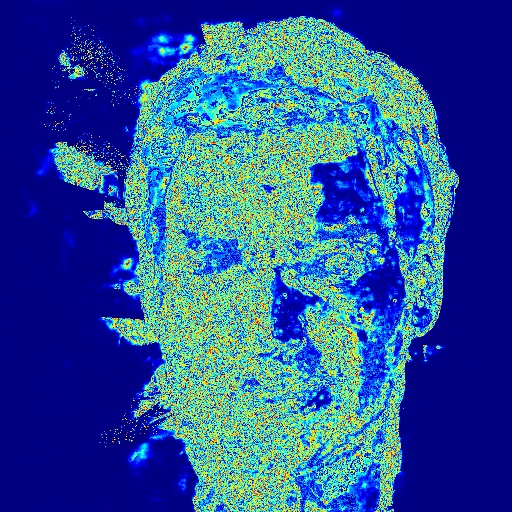} &
    \includegraphics[width=0.16\linewidth]{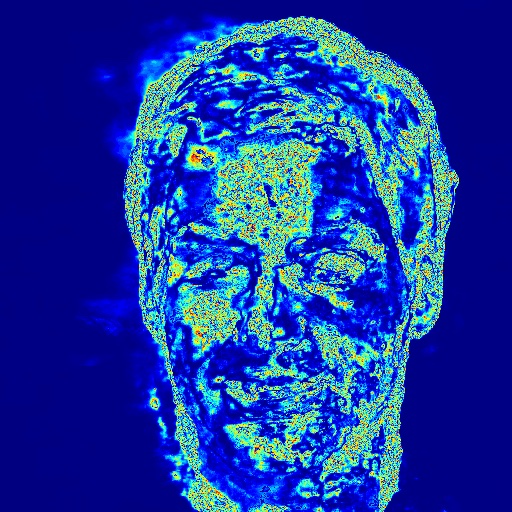} &
    \includegraphics[width=0.16\linewidth]{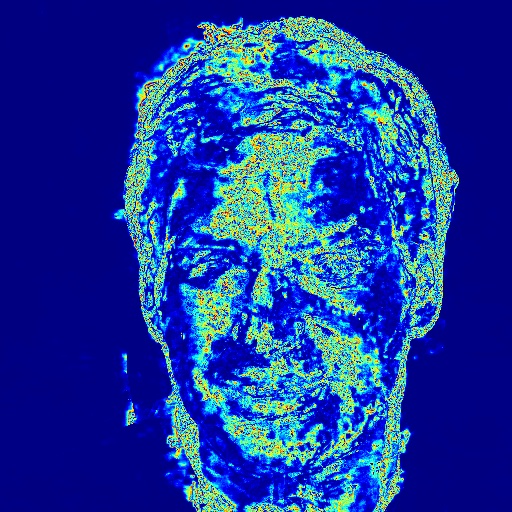} &
    \includegraphics[width=0.16\linewidth]{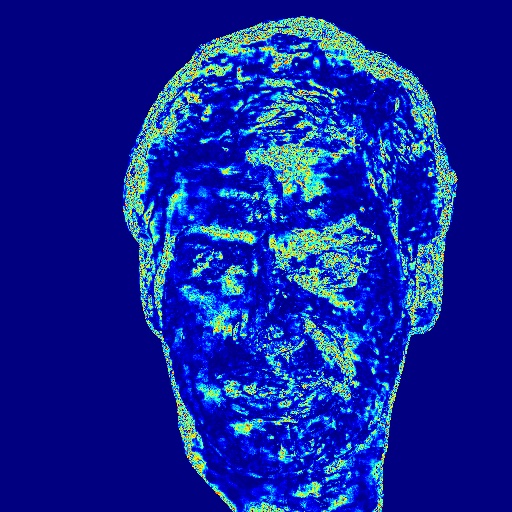} &
    
      \\
   \includegraphics[width=0.16\linewidth]{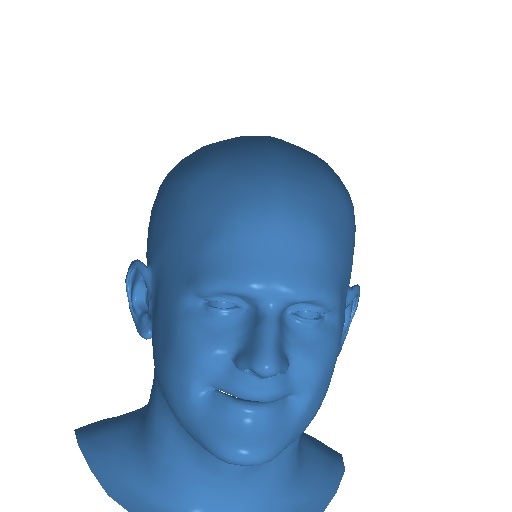} &
    \includegraphics[width=0.16\linewidth]{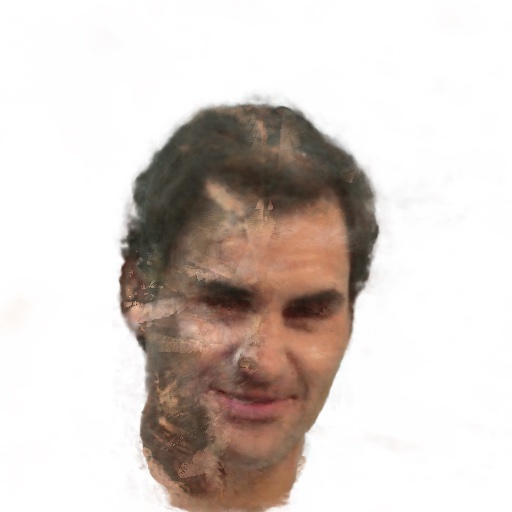} &
    \includegraphics[width=0.16\linewidth]{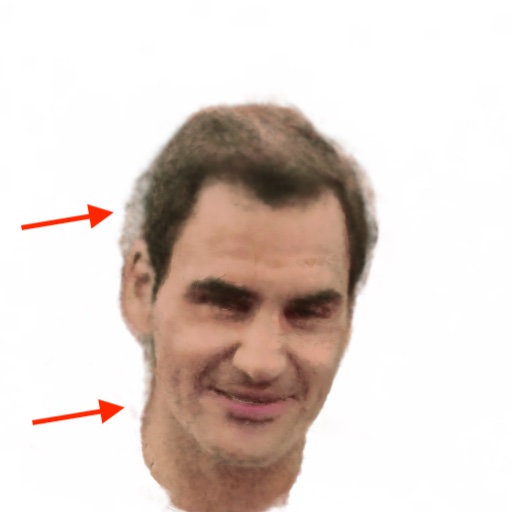} &
    \includegraphics[width=0.16\linewidth]{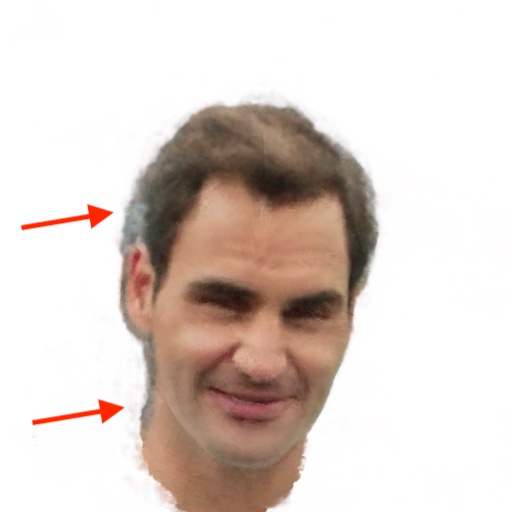} &
    \includegraphics[width=0.16\linewidth]{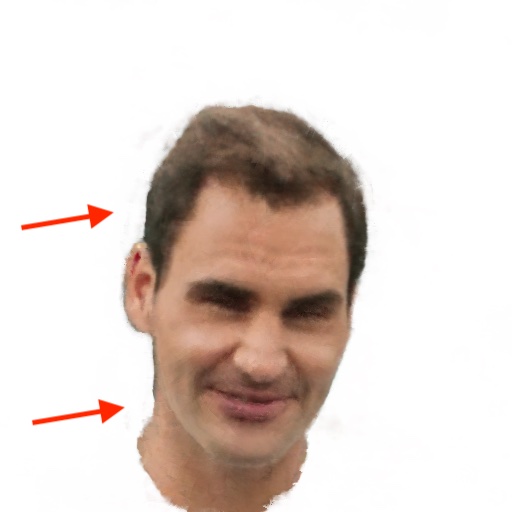} &
    \includegraphics[width=0.16\linewidth]{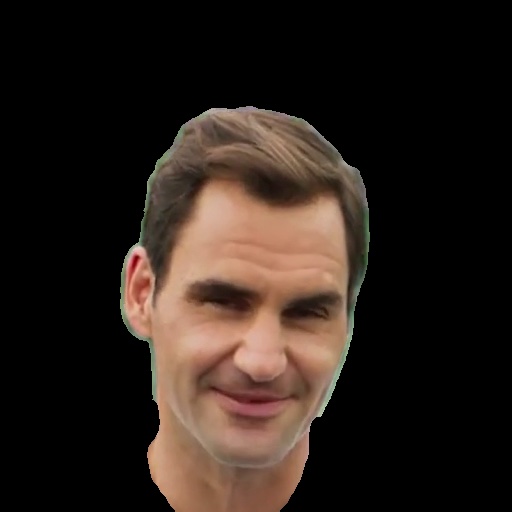} 
    \\
        &
    \includegraphics[width=0.16\linewidth]{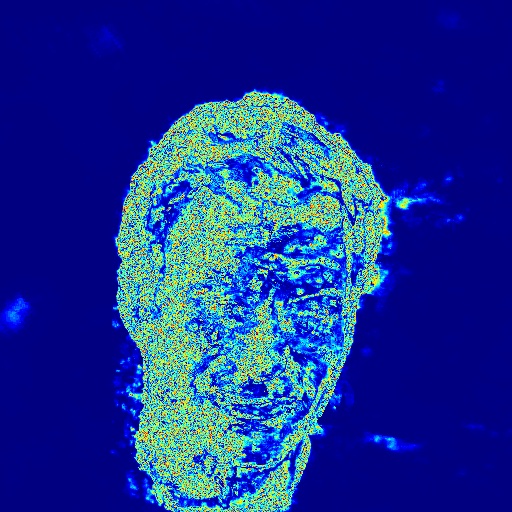} &
    \includegraphics[width=0.16\linewidth]{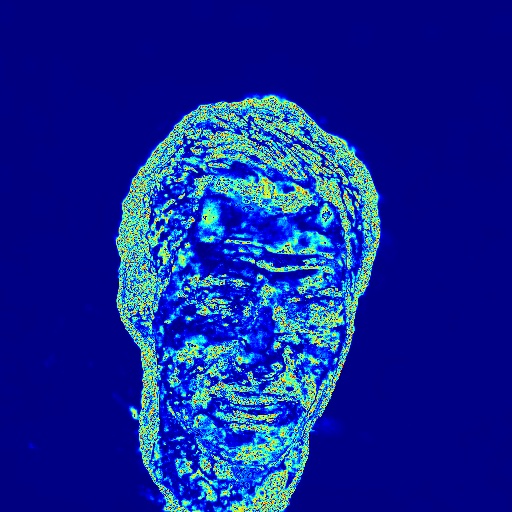} &
    \includegraphics[width=0.16\linewidth]{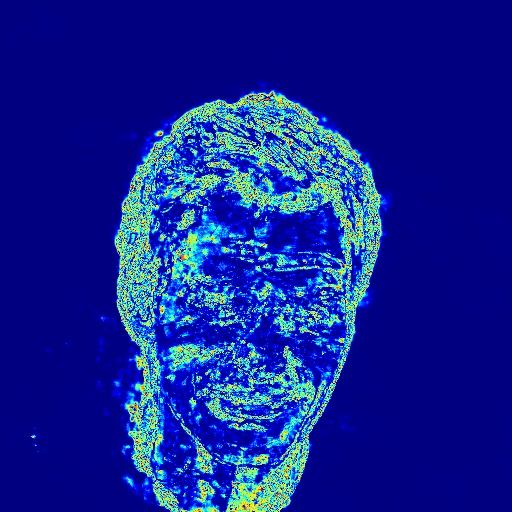} &
    \includegraphics[width=0.16\linewidth]{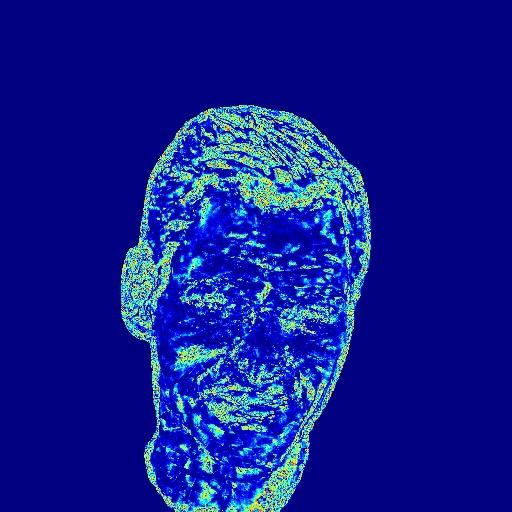} &
    
    \\
    \midrule
     (a) Target & (b) Insta \cite{zielonka2023instant} & (c) PAV w/o & (d) PAV w/o & (e) PAV & (f) GT\\ 
     Pose/Exp. &  & LNF & $\Delta_{\sigma}$ & Full & Image\\
  \end{tabular}
}
\caption{Qualitative Ablation Study on VidCol Dataset. This figure shows synthesized images and pixel error maps against ground-truth. Please note that brighter pixel colors denote higher pixel errors. Without appearance embedding with latent neural features and appearance-conditioned density offset (a), the model fails to learn appearance-based distinct details. The latent neural features (LNF) (c) resolve the issue of high-fidelity rendering of texture (\textcolor{blue}{blue arrow}). The appearance-conditioned density offset field (d) allows the model to learn more accurate geometry and texture that corresponds to the face, hair, and neck regions (\textcolor{red}{red arrow})}
\label{fig:ablation_study}
\end{figure*}

\begin{table}[h]
    \centering
    \setlength{\tabcolsep}{2mm}{
    \renewcommand\arraystretch{0.99}
    \resizebox{0.9\linewidth}{!}{
    \begin{tabular}{lccccc} 
    \toprule
    \toprule
    & \bf {PSNR\;$\uparrow$} & \bf {SSIM\;$\uparrow$} & \bf {DISTS\;$\downarrow$} & \bf {LPIPS\;$\downarrow$} & \bf {FID\;$\downarrow$}  \\  
    \midrule
    \midrule
    \bf Insta \cite{zielonka2023instant} & 21.03 & 0.849 & 0.185 & 0.169 & 104.28  \\
    \bf PAV without LNF  & 24.19 & 0.900 & 0.145 & 0.119 & 54.39  \\
    \bf PAV without $\Delta_{\sigma}$  & 24.14 & 0.906 & 0.134 & 0.108 & 47.71  \\
    \bf PAV (Full Model) & \bf 25.15 & \bf 0.910 & \bf 0.133 & \bf 0.107 & \bf 46.67  \\
    \midrule
    \bottomrule
    \end{tabular}}}
    \caption{Ablation Study: We evaluate the impact of different components of the proposed method. See Fig. \ref{fig:ablation_study} for qualitative comparison.} 
    \label{tab:ablation_study}
\end{table}

\subsection{Ablation Study}

We evaluate different design choices of PAV in Tab. \ref{tab:ablation_study} and Fig. \ref{fig:ablation_study}. In this study, we use protocol-a of the VidCol dataset for all experiments. We first report the results of the final model which includes \textit{latent neural features (LNF)} (Sec. \ref{sec:app_emb}) and density offset $\Delta_{\sigma}$ (Sec. \ref{sec:app_cond_nerf}). The full model achieves PSNR score of $25.15$ and SSIM score of $0.91$ for image synthesis. If we remove the density offset from our model while employing appearance-conditioned radiance estimation, PSNR and SSIM scores for image synthesis decrease to $24.14$ and $0.90$, respectively. This indicates that without appearance-conditioned density offset $\Delta_{\sigma}$, density estimation MLP ( $\mathcal{F}_{density}$ in Sec. \ref{sec:app_cond_nerf} ) can not learn to synthesize appearance-specific geometry and texture details, thus rendering fails there as well. Even though the model w/o $\Delta_{\sigma}$ synthesise images that look on par with ground truth in terms of texture, it generates severe artifacts which can be seen in pixel error maps. An example of this behavior can also be seen in Fig. \ref{fig:ablation_study} (w/o $\Delta_{\sigma}$). Rendering failures on face, head, and neck regions are indicated by \textcolor{red}{red arrow} and error map. 

Next, we evaluate the impact of \textit{latent neural features} (LNF) as an \textit{appearance embedding} on PAV. In this experiment, to condition the canonical MLPs in PAV, inspired by "Nerf In-the-wild" \cite{martin2021nerf,weng2023personnerf}, we adopt Generative Latent Optimization (GLO) for appearance embedding. Specifically, this approach defines a single real-valued appearance embedding vector $l$ for each appearance in the dataset. To this end, we employ a pre-trained encoder from in-the-wild face detector RetinaFace \cite{deng2019retinaface} for GLO. During training, if we replace  $\mathbf{Z}{=}\{Z_\mathrm{i}\}_{i=1}^N, Z_{i} \in \mathbb{R}^{512 \times 512 \times 16}$ with $\mathbf{L}{=}\{l_\mathrm{i}\}_{i=1}^N, l_{i} \in \mathbb{R}^{512 \times 1}$ for $N$ appearances of each subject, and pass it to appearance conditioned NeRF (See Sec. \ref{sec:app_cond_nerf}), PSNR and SSIM scores for image synthesis decrease to 24.19 and 0.90, respectively. This is also evident from the qualitative results shown in Fig. \ref{fig:ablation_study} (w/o LNF), indicating that latent neural features help in improved image synthesis quality with their local features attached to 3D geometry. Rendering failures on head, neck, and skin texture are shown by \textcolor{red}{red arrow}, \textcolor{blue}{blue arrow}  and error maps in Fig. \ref{fig:ablation_study}. Finally, as discussed in Sec. \ref{sec:method}, we keep the network design as we explain after the ablation study. 



\subsection{Comparison}
\begin{figure*}[t]
  \centering
  \small
    \setlength{\tabcolsep}{1pt}
    \renewcommand{\arraystretch}{0.0}
\scalebox{0.95}{
  \begin{tabular}{cccccccc}
    \includegraphics[width=0.125\linewidth]{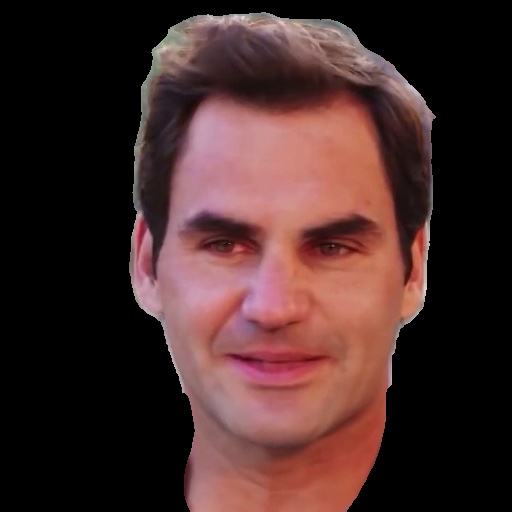} &
    \includegraphics[width=0.125\linewidth]{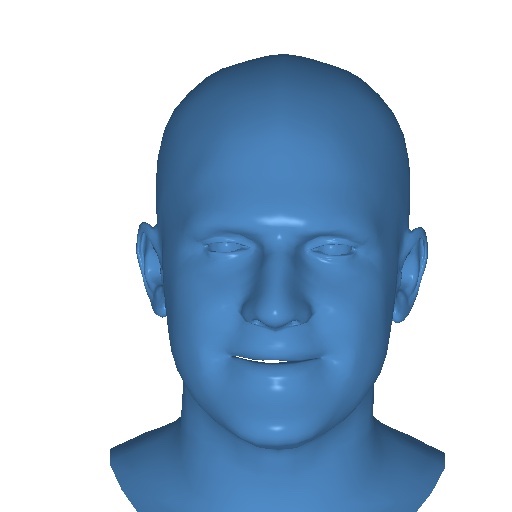} &
    \includegraphics[width=0.125\linewidth]{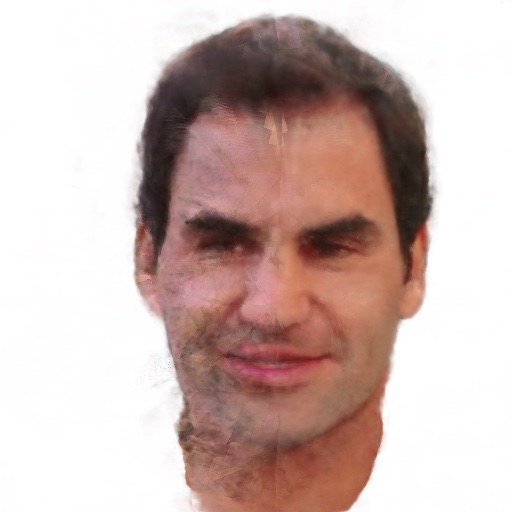} &
    \includegraphics[width=0.125\linewidth]{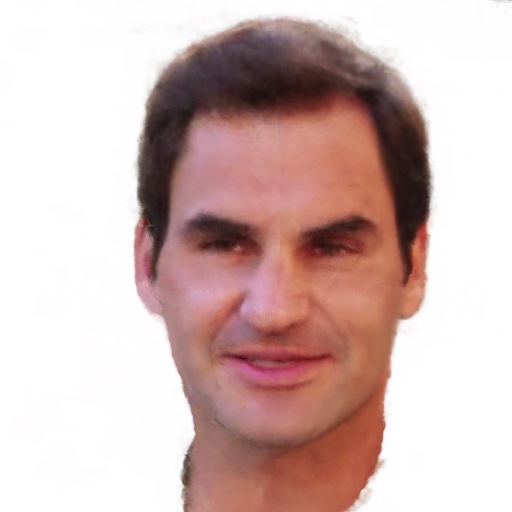} &
    \includegraphics[width=0.125\linewidth]{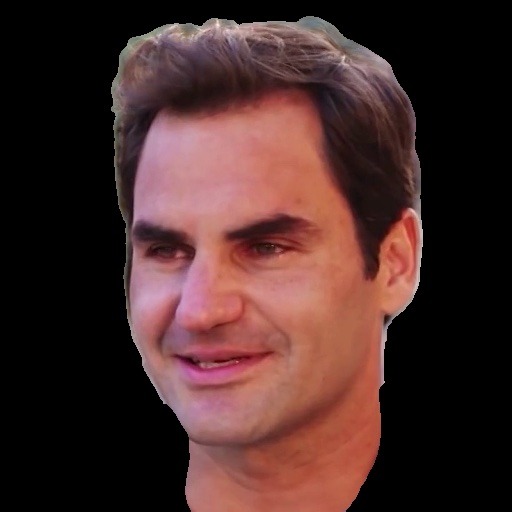} &
    \includegraphics[width=0.125\linewidth]{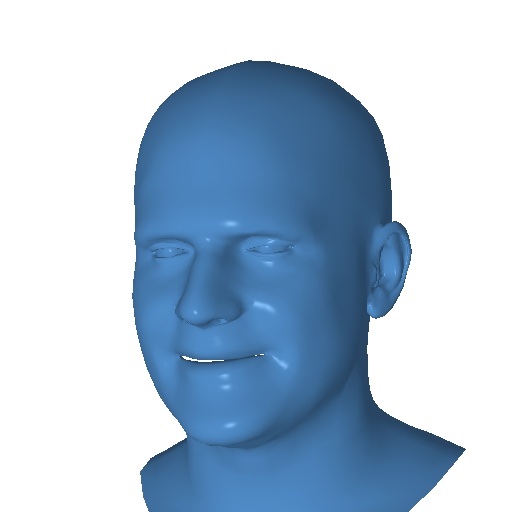} &
    \includegraphics[width=0.125\linewidth]{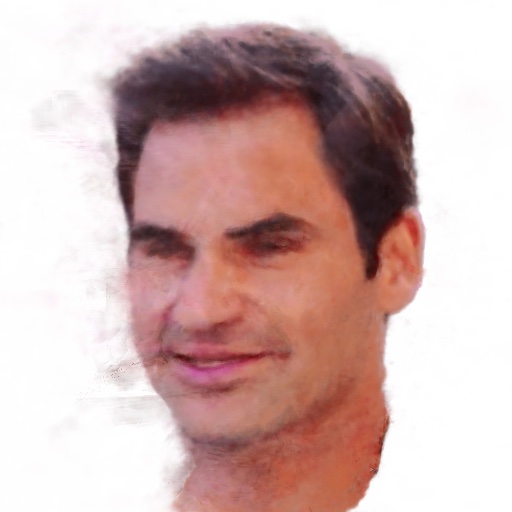} &
    \includegraphics[width=0.125\linewidth]{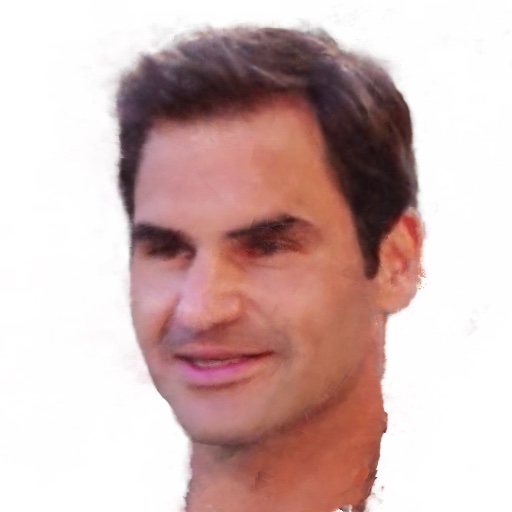}  \\
    \includegraphics[width=0.125\linewidth]{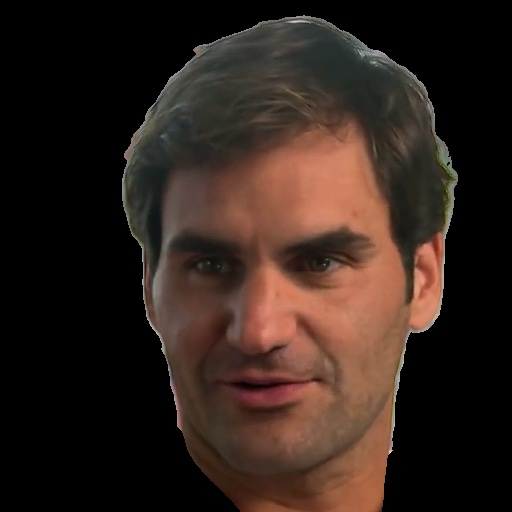} &
    \includegraphics[width=0.125\linewidth]{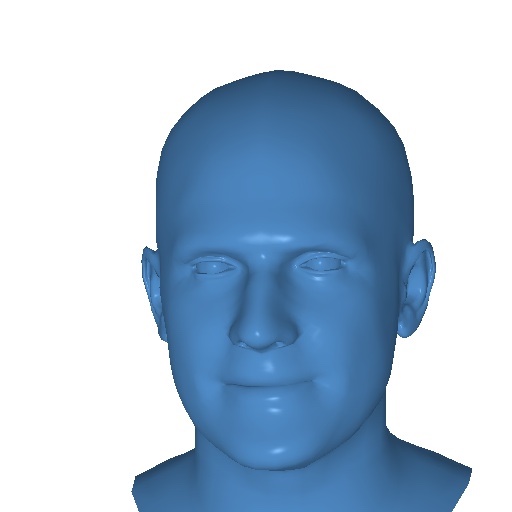} &
    \includegraphics[width=0.125\linewidth]{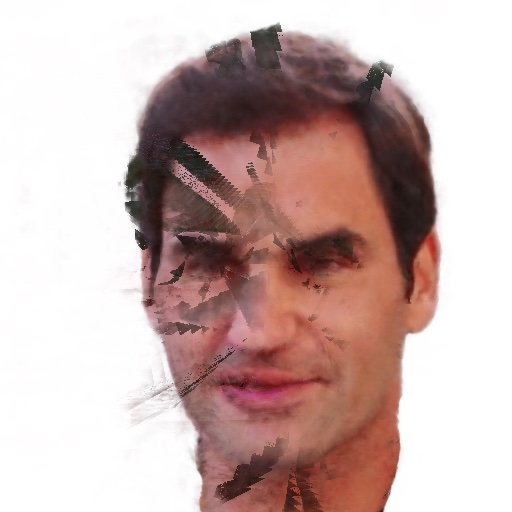} &
    \includegraphics[width=0.125\linewidth]{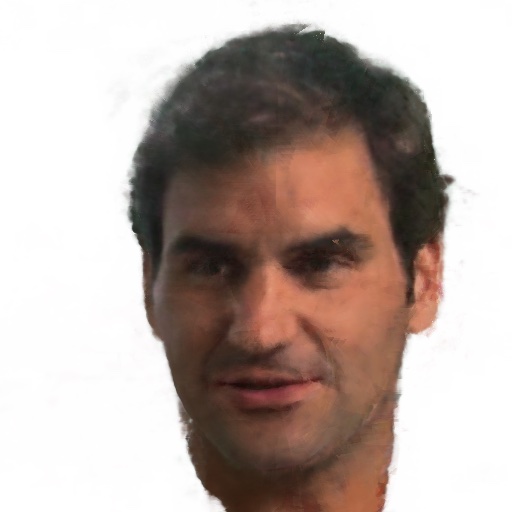} &
    \includegraphics[width=0.125\linewidth]{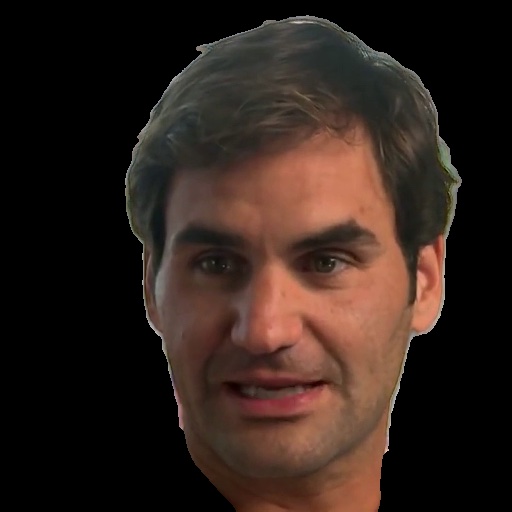} &
    \includegraphics[width=0.125\linewidth]{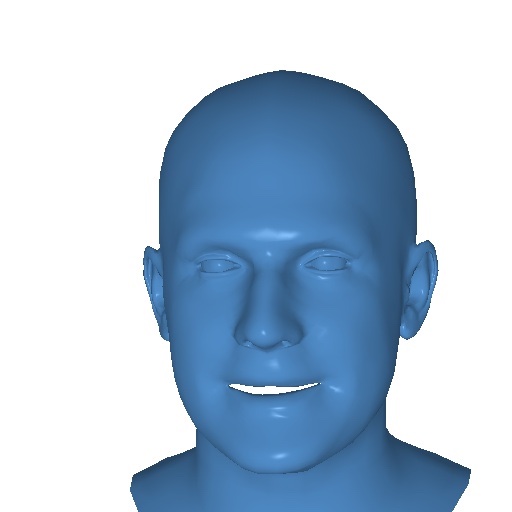} &
    \includegraphics[width=0.125\linewidth]{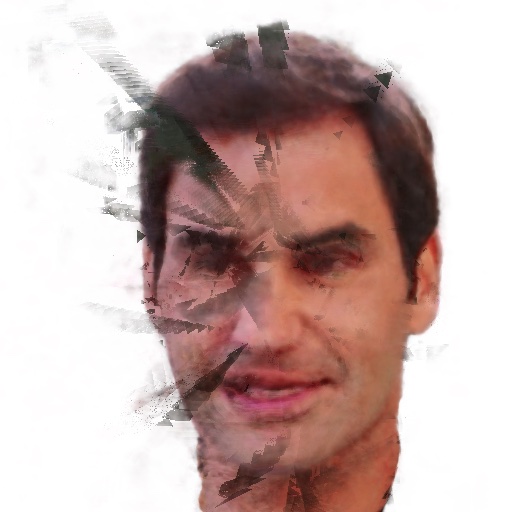} &
    \includegraphics[width=0.125\linewidth]{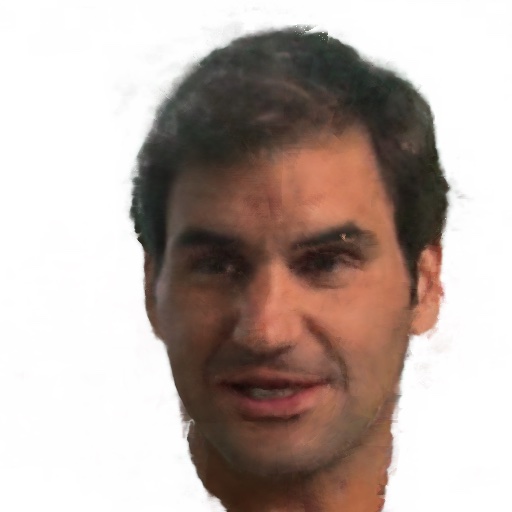} \\
    \includegraphics[width=0.125\linewidth]{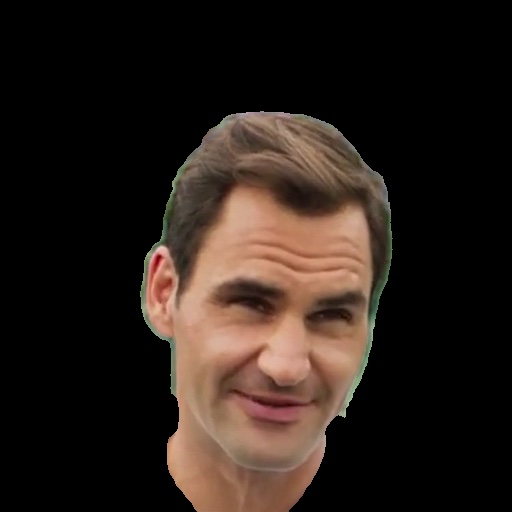} &
    \includegraphics[width=0.125\linewidth]{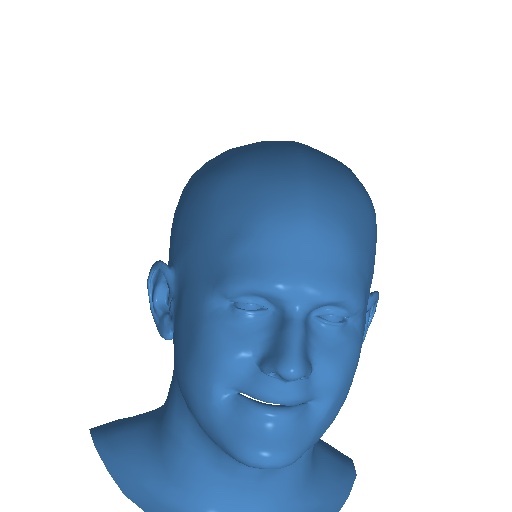} &
    \includegraphics[width=0.125\linewidth]{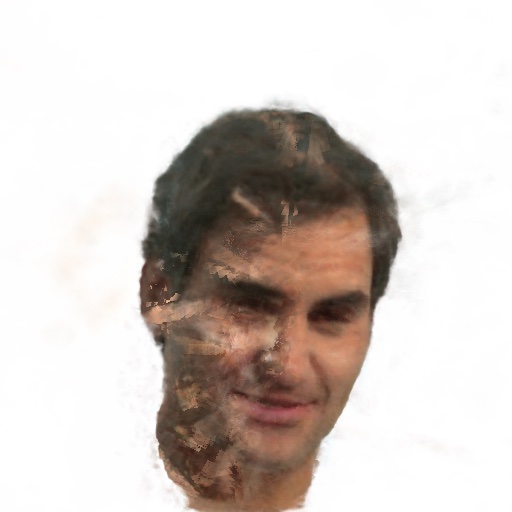} &
    \includegraphics[width=0.125\linewidth]{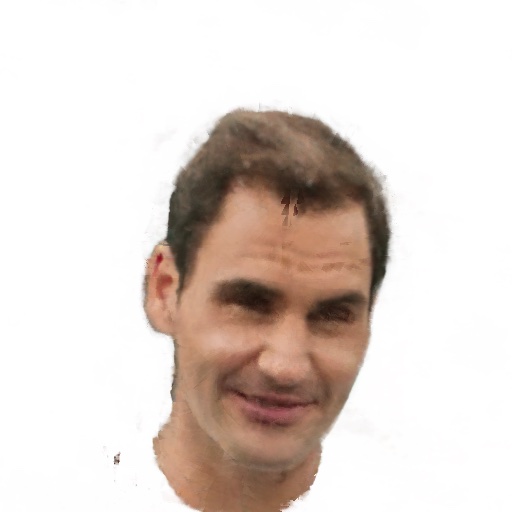} &
    \includegraphics[width=0.125\linewidth]{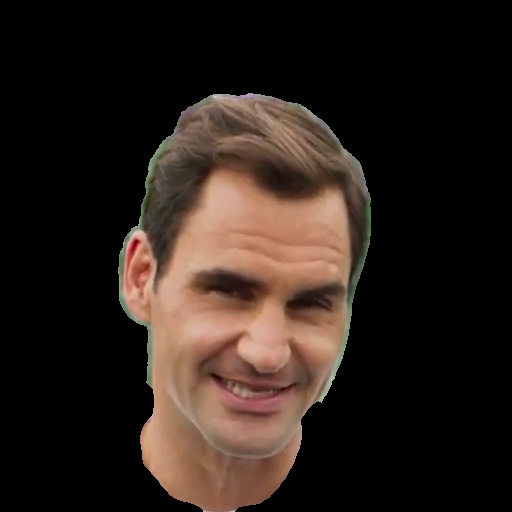} &
    \includegraphics[width=0.125\linewidth]{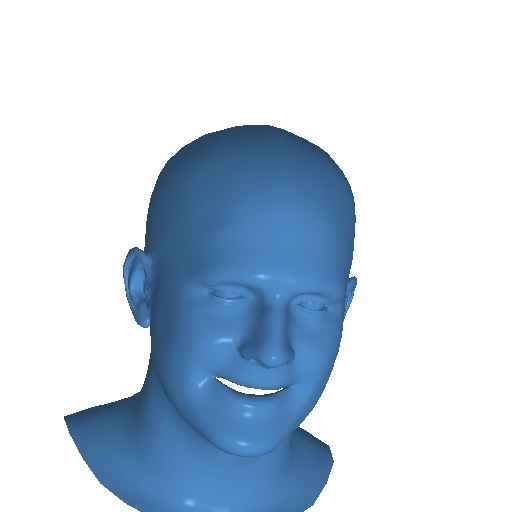} &
    \includegraphics[width=0.125\linewidth]{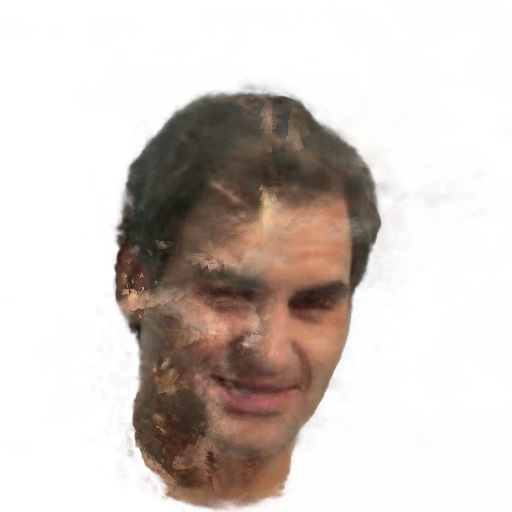} &
    \includegraphics[width=0.125\linewidth]{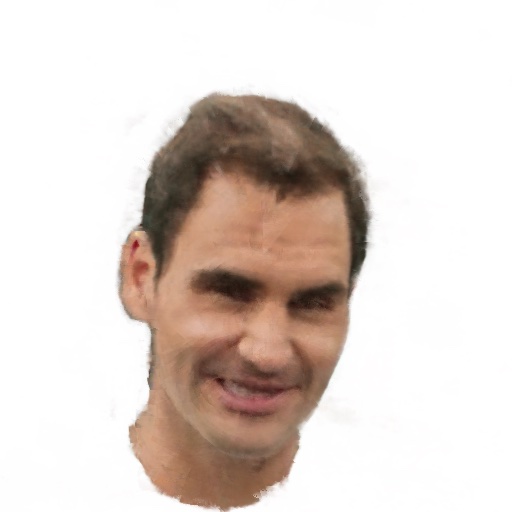} \\
    \midrule \\
    \includegraphics[width=0.125\linewidth]{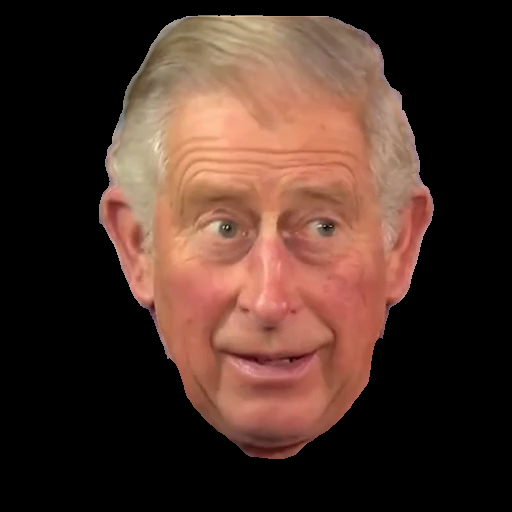} &
    \includegraphics[width=0.125\linewidth]{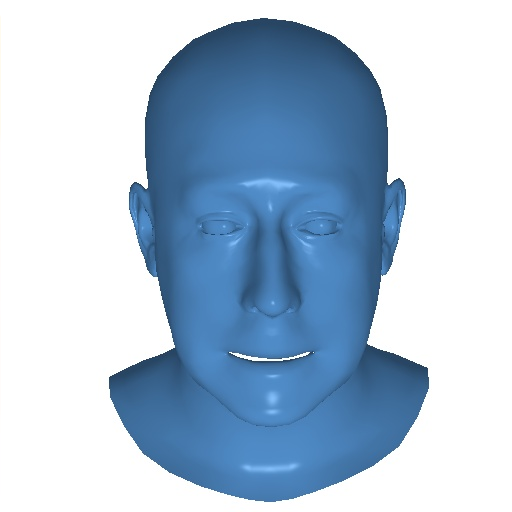} &
    \includegraphics[width=0.125\linewidth]{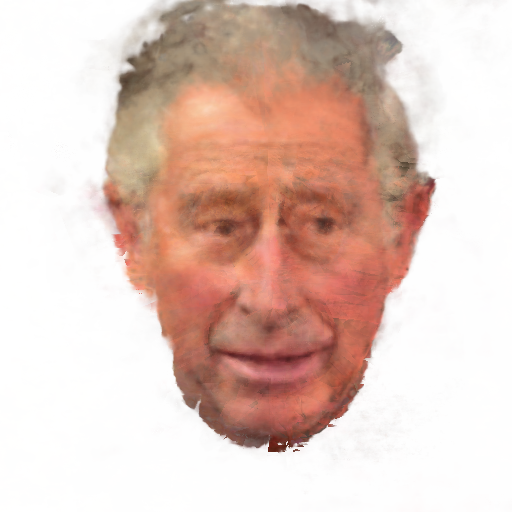} &
    \includegraphics[width=0.125\linewidth]{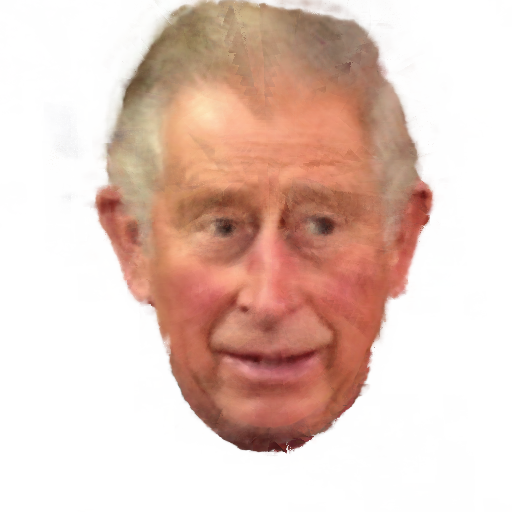} &
    \includegraphics[width=0.125\linewidth]{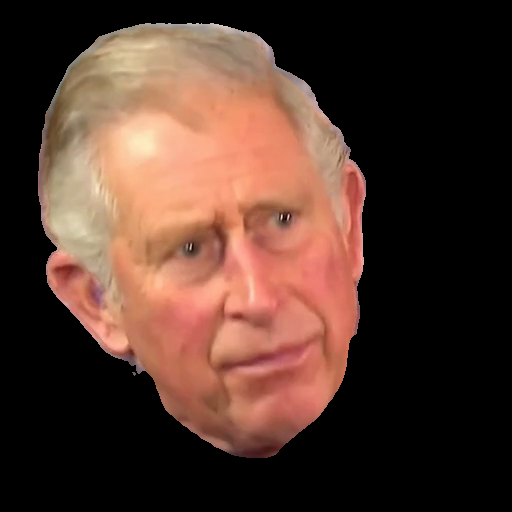} &
    \includegraphics[width=0.125\linewidth]{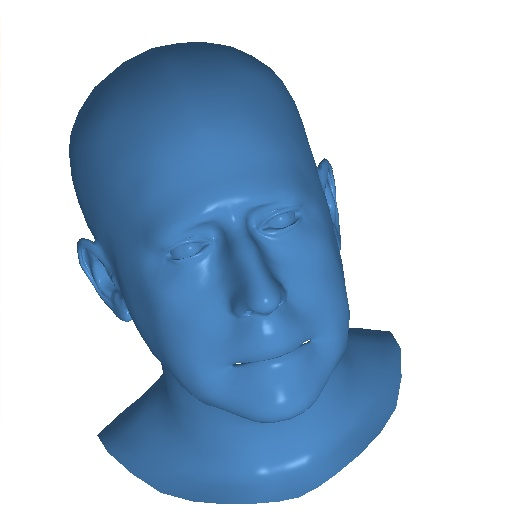} &
    \includegraphics[width=0.125\linewidth]{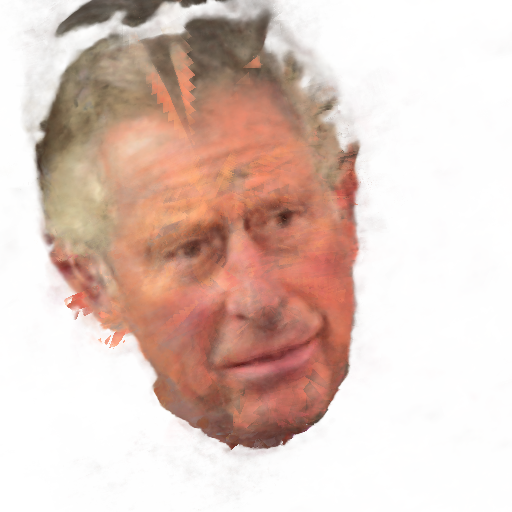} &
    \includegraphics[width=0.125\linewidth]{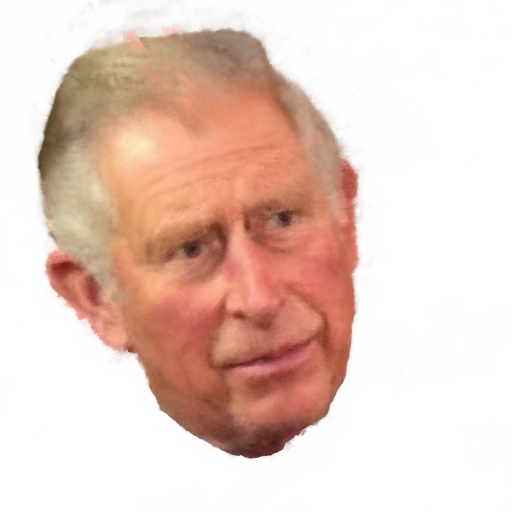} \\
    \includegraphics[width=0.125\linewidth]{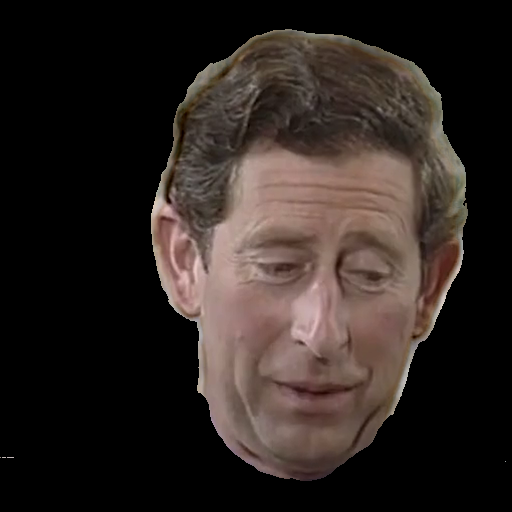} &
    \includegraphics[width=0.125\linewidth]{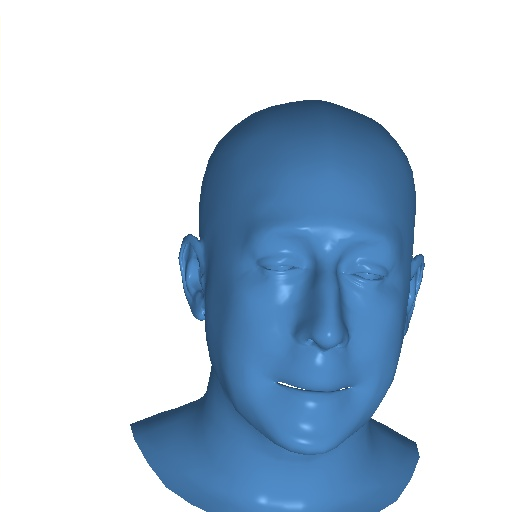} &
    \includegraphics[width=0.125\linewidth]{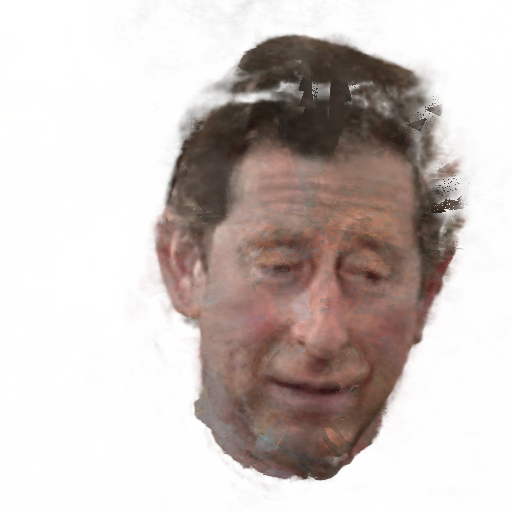} &
    \includegraphics[width=0.125\linewidth]{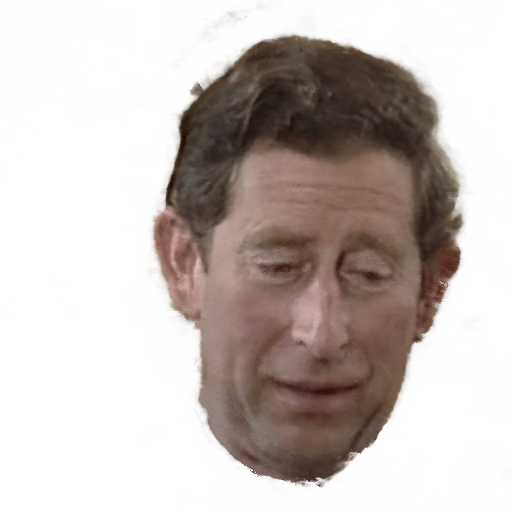} &
    \includegraphics[width=0.125\linewidth]{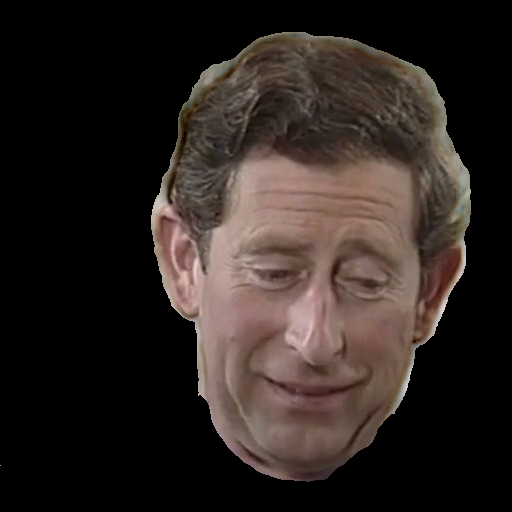} &
    \includegraphics[width=0.125\linewidth]{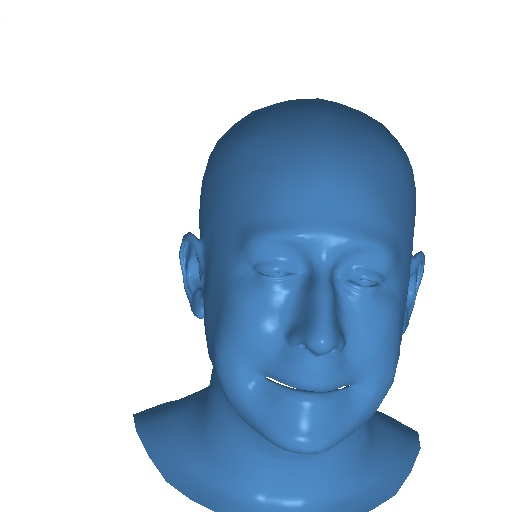} &
    \includegraphics[width=0.125\linewidth]{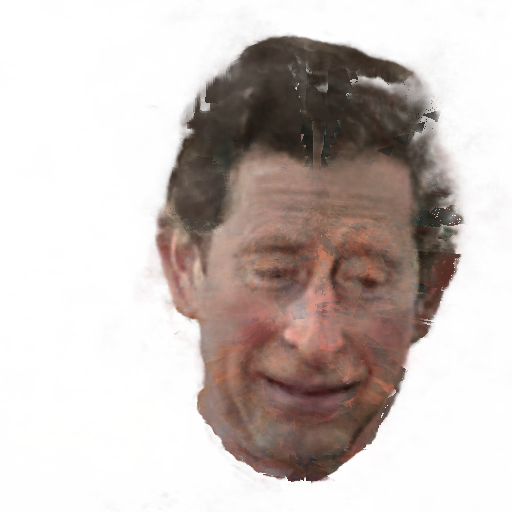} &
    \includegraphics[width=0.125\linewidth]{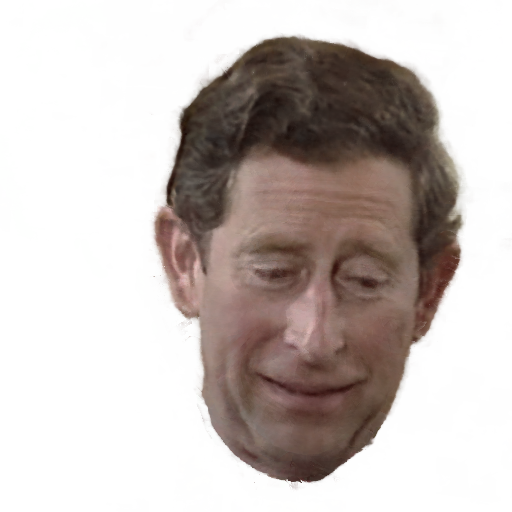} \\
    \includegraphics[width=0.125\linewidth]{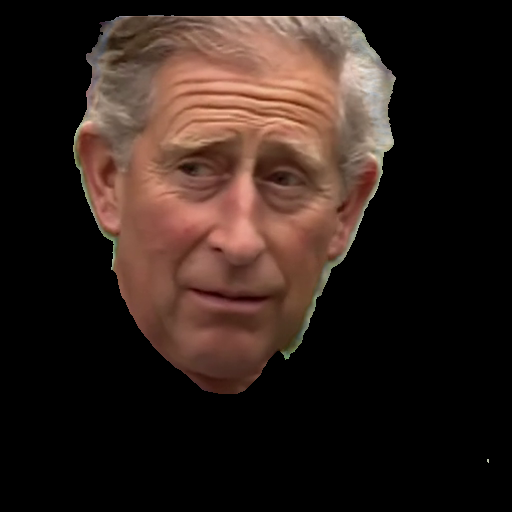} &
    \includegraphics[width=0.125\linewidth]{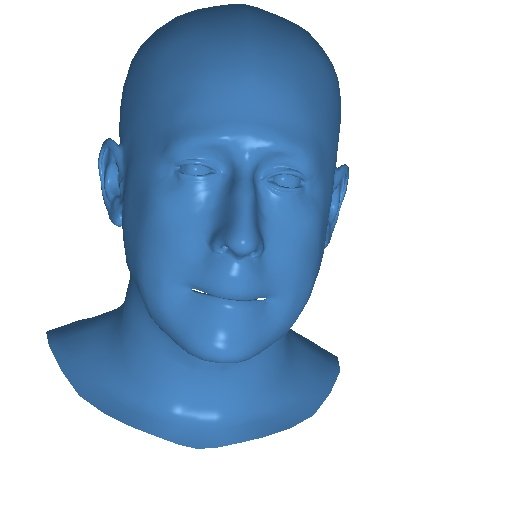} &
    \includegraphics[width=0.125\linewidth]{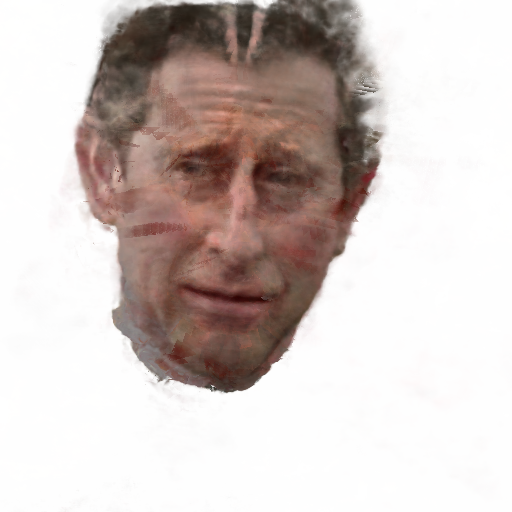} &
    \includegraphics[width=0.125\linewidth]{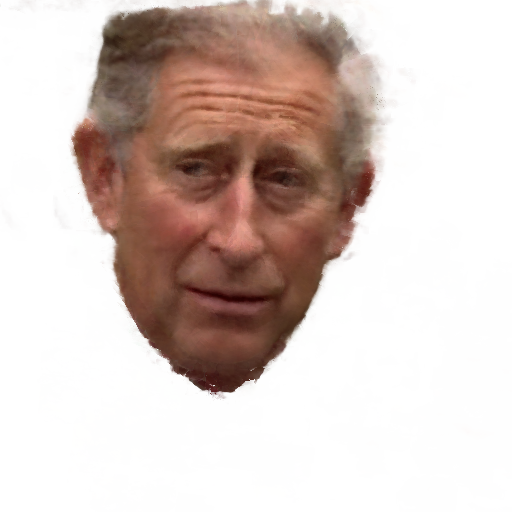} &
    \includegraphics[width=0.125\linewidth]{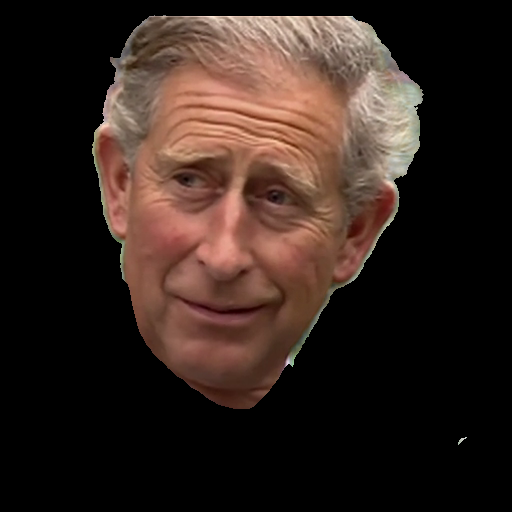} &
    \includegraphics[width=0.125\linewidth]{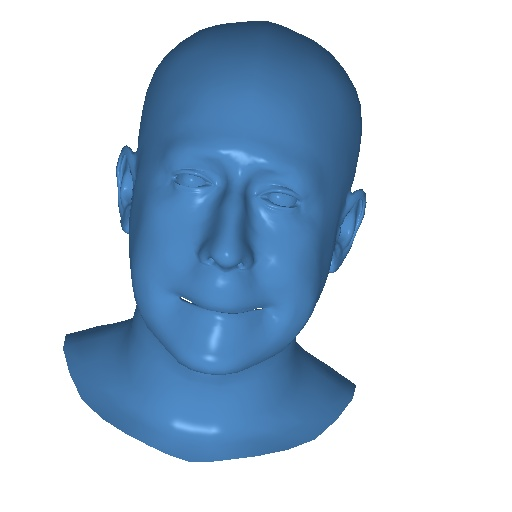} &
    \includegraphics[width=0.125\linewidth]{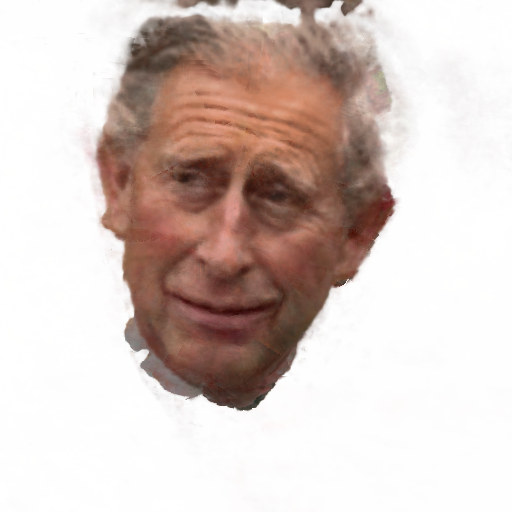} &
    \includegraphics[width=0.125\linewidth]{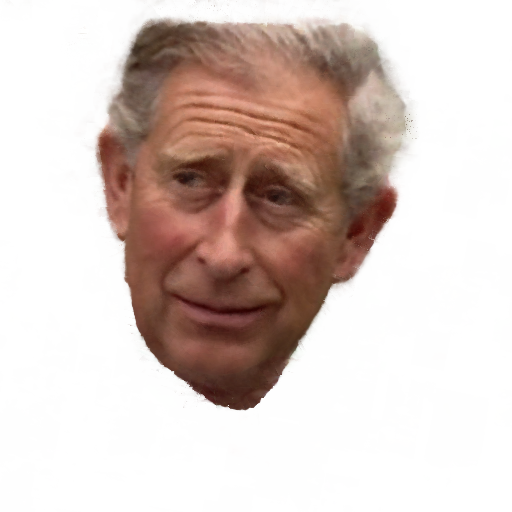} \\
    \midrule
    GT & Target & Insta \cite{zielonka2023instant} & PAV (Ours) & GT & Target & Insta \cite{zielonka2023instant} & PAV (Ours)\\ 
    Image & Pose/Exp. & Render & Render & Image & Pose/Exp.
    & Render & Render\\
  \end{tabular}
}
\caption{Qualitative comparison of our method (PAV) against baseline method (Insta \cite{zielonka2023instant}. Columns \{2,6\} show target head poses and facial expressions for synthesized images at columns \{3,4,7,8\}. }
\label{fig:comparison_baseline}
\end{figure*}

Since our method is the first work that learns dynamic deformable head NeRF from multiple appearances of the same subjects using a unified network, we ourselves design some baseline comparisons in this section. First, we compare our method against Insta \cite{zielonka2023instant} using protocol-a (Sec. \ref{sec:exp_dataset}) of the VidCol dataset. Both methods are trained using the frames sampled from all appearances of the same subject. For testing, both methods synthesize images for target head pose and expression. The results are summarized in Tab. \ref{tab:baselines_and_sota} and Fig. \ref{fig:comparison_baseline}. For protocol-a, our method clearly outperforms the baseline for final image synthesis results across all metrics. It should be noted that PAV framework jointly learns all different appearances via one-time training using a unified network and it is able to preserve appearance-specific details for novel poses and expressions. On the other hand, as shown in Fig. \ref{fig:comparison_baseline}, Insta can not disentangle appearance from the neural radiance field, which results in a mixture of appearances in synthesized images and inaccurate geometry predictions. This is the limitation of Insta we are addressing in the paper. 


\begin{table*}[h]
    \centering
    \setlength{\tabcolsep}{2mm}{
    \vspace{2pt}
    \renewcommand\arraystretch{0.99}
    \resizebox{1.0\linewidth}{!}{
    \begin{tabular}{lccccc|ccccc} 
    \toprule
    \multirow{2}{*}{Method}  & 
    \multicolumn{5}{c}{\textbf{Video Collection 1}}  & \multicolumn{5}{c}{\textbf{Video Collection 2}}\\ 
    & {PSNR\;$\uparrow$} &  {SSIM\;$\uparrow$} & {DISTS\;$\downarrow$} & {LPIPS\;$\downarrow$} & {FID\;$\downarrow$} & {PSNR\;$\uparrow$} &  {SSIM\;$\uparrow$} & {DISTS\;$\downarrow$} & {LPIPS\;$\downarrow$} & {FID\;$\downarrow$} \\  
    \midrule
    \multicolumn{11}{c}{\bf{Protocol (a):} Novel Head Pose and Expression Synthesis} \\ 
    \midrule
    Insta \cite{zielonka2023instant} & 21.03 & 0.84 & 0.185 & 0.169 & 104.28 & 23.67 & 0.86 & 0.187 & 0.163 & 98.87\\
    PAV (Ours) &  \bf 25.15  &  \bf 0.91  &  \bf 0.133  & \bf 0.107  & \bf 46.67  & \bf 27.09  & \bf 0.89  & \bf 0.144 & \bf 0.120 & \bf 53.91 \\
    \midrule
    \bottomrule
    \end{tabular}}}
    \caption{Comparison of PAV against Insta \cite{zielonka2023instant}. See Fig. \ref{fig:comparison_baseline} for qualitative comparison. Video Collection 1 and 2 refer to \textit{Federer} and \textit{King Charles} splits of \textit{VidCol} dataset, respectively.} 
    \label{tab:baselines_and_sota}
\end{table*}

\begin{figure}[h]
  \centering
  \small
  \setlength{\tabcolsep}{0pt}
  \renewcommand{\arraystretch}{0.0}
  \scalebox{0.7}{
  \begin{tabular}{cccc}
  {\includegraphics[width=0.25\linewidth]{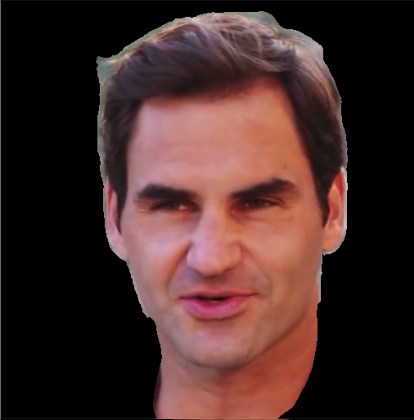}} & 
  {\includegraphics[width=0.25\linewidth]{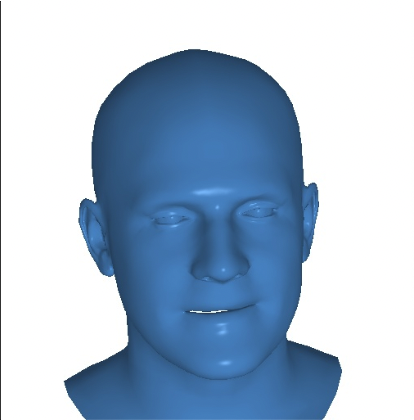}} & 
  {\includegraphics[width=0.25\linewidth]{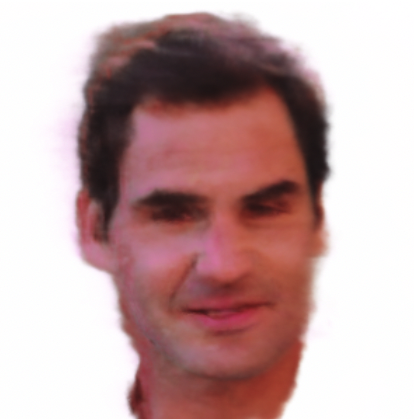}} &
  {\includegraphics[width=0.25\linewidth]{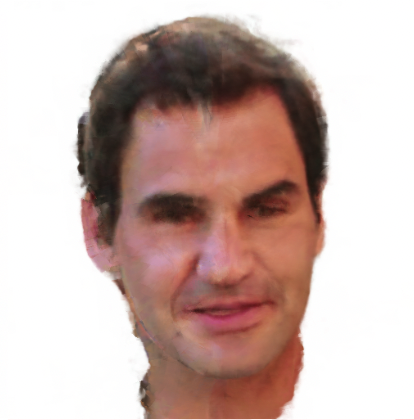}} \\
  {\includegraphics[width=0.25\linewidth]{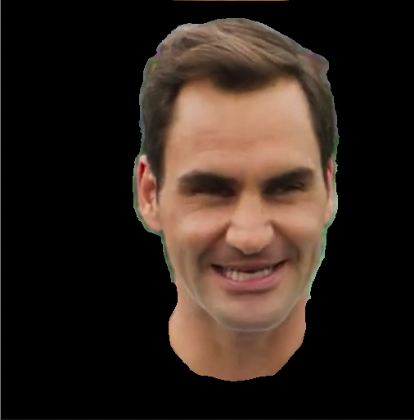}} & 
  {\includegraphics[width=0.25\linewidth]{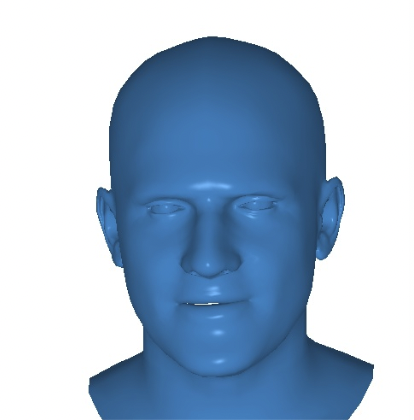}} & 
  {\includegraphics[width=0.25\linewidth]{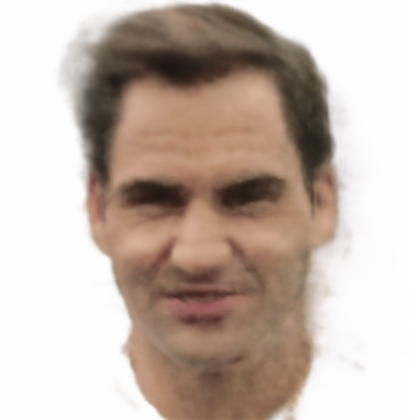}} &
  {\includegraphics[width=0.25\linewidth]{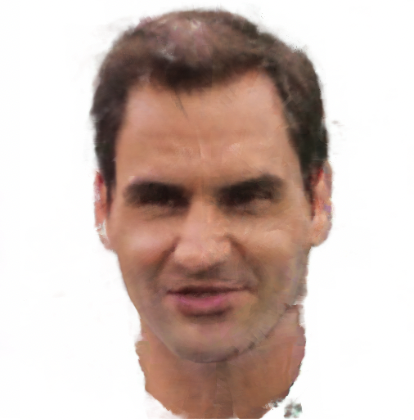}}
  \\
  \midrule
  Reference & Target Pose/Exp. & Insta \cite{zielonka2023instant} & PAV
  \end{tabular}
  }
  \caption{Qualitative comparison of PAV against single-appearance traines Insta \cite{zielonka2023instant}. Please note that the reference image is for photoconsistency check only since there is no ground truth image for that pose/expression
  }
  \label{fig:comparison_single_nerf}
\end{figure}

Second, we train separate Insta \cite{zielonka2023instant} models for each appearance of the same subject using protocol-b while training a single unified model for our method using the frames sampled from all appearances of the same subject. For testing, we use the same test data from protocol-b for both PAV and Insta \cite{zielonka2023instant} and synthesize images for novel poses and expressions. Note that, we added more novel expressions and poses in protocol-b (Sec. \ref{sec:exp_dataset}) to test the generalization of our method under a unified training schema, which results in increasing FID score for both methods in Tab. \ref{tab:comparison_single_apperance_nerf}. Even though the results show that artifacts (in Fig. \ref{fig:comparison_baseline}) of the Insta renders are decreased in Fig. \ref{fig:comparison_single_nerf}, our method demonstrates better-rendering results (Fig. \ref{fig:comparison_single_nerf}) and lower FID score (Tab. \ref{tab:comparison_single_apperance_nerf}). This indicates that PAV's single unified model helps in learning generic neural radiance field representation across all appearances of the same subject while preserving appearance-based details. Please note that we are demonstrating Insta's performance under a challenging pose/expression in Fig. \ref{fig:comparison_single_nerf} which can also be seen in the original paper results under extreme poses/expressions. In Fig. \ref{fig:comparison_single_nerf}, the reference image is for photoconsistency check only since there is no ground truth image for that pose/expression. Finally, we provide additional qualitative results of avatars animated in the supplementary video. We also compared PAV against Insta for model efficiency using protocol-b dataset and recorded values in Tab. \ref{tab:comparison_single_apperance_nerf}. Although we train our model with multiple appearances jointly, PAV is more efficient than Insta in terms of memory footprint and training time


\begin{table}[h]
    \centering
    \setlength{\tabcolsep}{2mm}{
    \renewcommand\arraystretch{0.99}
    \resizebox{1.0\linewidth}{!}{
    \begin{tabular}{lcccc} 
    \toprule
    \toprule
    & \bf {Train Time} & \bf { Number of Param.} & \bf {Model} &  \bf {FID\;$\downarrow$}  \\ 
    & \bf {Minutes} & \bf {Million} & \bf {Size (MB)} & \bf {}  \\ 
    \midrule
    \midrule
    \bf Multiple Insta \cite{zielonka2023instant} & 135 & 79.7 & 1932 & 109.22  \\
    \bf PAV (Ours)  & \bf 45 & \bf 26.8 & \bf 650  & \bf 91.01  \\
    \midrule
    \bottomrule
    \end{tabular}}}
    \caption{Efficiency and performance comparison between PAV and Insta \cite{zielonka2023instant} trained for each appearance (Single Appearance Training) in dataset collection 1. FID score is computed from synthetized images under novel head poses and expressions where there are no existing grount-truth images.} 
    \label{tab:comparison_single_apperance_nerf}
\end{table}












\section{Discussion}

We presented PAV which is a novel method for learning personalized head avatars from monocular and in-the-wild video collections of a person. We demonstrated that PAV can model a 3D head avatar from an unstructured video collection of a subject in a single unified network while disentangling face appearances from the radiance field. PAV can be trained simultaneously for multiple appearances of the same subject, allowing novel pose and expression generalization and leading to a superior rendering quality of our method compared to the baseline. 



\begin{figure}[h]
  \centering
  \small
  \setlength{\tabcolsep}{0pt}
  \renewcommand{\arraystretch}{0.0}
  \scalebox{0.7}{
  \begin{tabular}{ccc}
  {\includegraphics[width=0.25\linewidth]{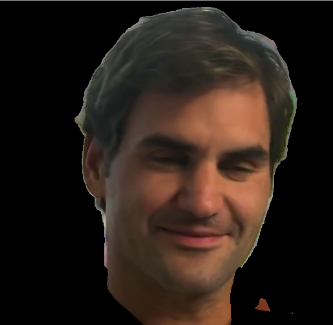}} & 
  {\includegraphics[width=0.25\linewidth]{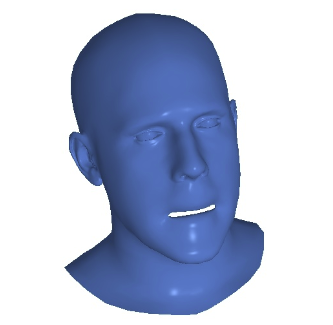}} & 
  {\includegraphics[width=0.25\linewidth]{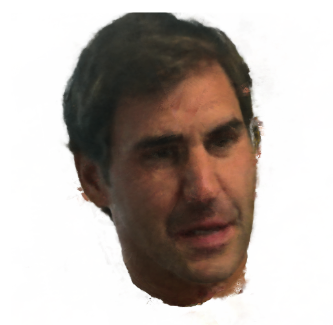}} \\
  {\includegraphics[width=0.25\linewidth]{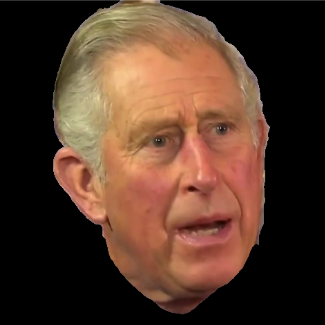}} & 
  {\includegraphics[width=0.25\linewidth]{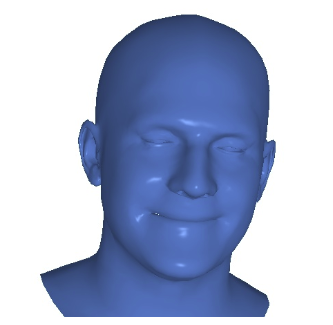}} & 
  {\includegraphics[width=0.25\linewidth]{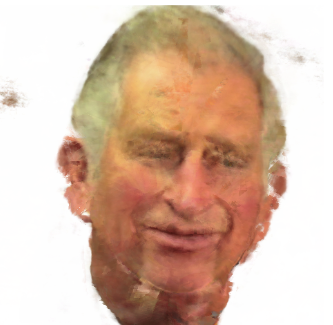}}
  \\
  \midrule
  Reference Subject & Target Pose/Exp. &  Ours (PAV)
  \end{tabular}
  }
  \caption{Multi-identity and multi-appearance extension of PAV. The learned appearance (middle) rendered in a new viewpoint (left) does not match the visual quality of input images (right). Please note that the reference image is for photoconsistency check only since there is no ground truth image for that pose/expression.
  }
  \label{fig:multi_identity}
\end{figure}

\paragraph{Limitation for Multi-Identity.}


We extended our method for multiple identities and multiple appearances for each identity. To this end, we train a single unified PAV model using frames smapled from all appearances of \textit{Federer} and \textit{King} subjects from \textit{VidCol} dataset with shared canonical space. We evaluate novel head pose and facial expression rendering of this model and Fig. \ref{fig:multi_identity} illustrates the results. Even though this model produces reasonable results, synthesized images demonstrate a lack of subject-specific shape and appearance details with rendering artifacts. In the paper, we focus on multiple appearances of a single subject. Scaling up this model to a wide range of subjects and exploring generalized dynamic deformable NeRF would be interesting future work.


\paragraph{Ethical Considerations.}

Although the rapid progress on digital human avatars facilitates numerous downstream applications, it also raises concerns about ethical misuse. On the synthesis side, it would be ideal to actively use watermarking technologies and avoid driving avatars with different identities. In this work, we aim to faithfully reproduce images of a person with the capability of rendering unseen poses and expressions and switching appearance within their own set of appearances. The work does not intend to create animations that didn't happen. 






%
%
\bibliographystyle{splncs04}
\bibliography{main}
\end{document}